\documentclass[leqno,11pt]{article}

\usepackage{subcaption}
\usepackage{xspace}
\usepackage{amsmath}
\usepackage{tikz-cd}
\usepackage{mathtools}
\usepackage{multicol}

\usepackage[ruled,linesnumbered,lined,commentsnumbered]{algorithm2e}

\usepackage[english]{babel}
\usepackage{amsmath,amsthm,amssymb,mathrsfs} 
\usepackage{ae,aecompl}
\usepackage{times}
\usepackage[pagebackref,hypertexnames=false]{hyperref}
\usepackage{fancybox,fancyhdr,graphics,epsfig}
\usepackage{textcomp}
\usepackage{authblk}

\newtheorem{lem}{Lemma}[section]

\newtheorem{remark}[lem]{Remark}

\theoremstyle{definition}
\newtheorem{definition}[lem]{Definition}

 \usepackage{amsmath,amsthm,bbm,amssymb}

\DeclareMathOperator{\PH}{PH}
\DeclareMathOperator{\PD}{PD}
\DeclareMathOperator{\Ker}{Ker}
\DeclareMathOperator{\Imm}{Im}
\DeclareMathOperator{\bth}{birth}
\DeclareMathOperator{\dth}{death}

\DeclareMathOperator{\low}{low}
\DeclareMathOperator{\rank}{rank} 
\DeclareMathOperator{\Int}{Int}

\DeclareMathOperator{\Cech}{\check{C}ech}
\DeclareMathOperator{\Rips}{Rips}

\DeclareMathOperator{\spn}{span}

\newcommand{\cech}{\v{C}ech\xspace}

\newcommand{\bcap}{\bigcap\limits}

\newcommand\R{{\mathbb{R}}}

\def\cX{{\cal{X}}}
\def\cY{{\cal{Y}}}
\def\cZ{{\cal{Z}}}

\begin{document}

\title{Cycle Registration in Persistent Homology with Applications in Topological Bootstrap}

\author[1]{Yohai Reani \thanks{syohai@campus.technion.ac.il}}
\author[1]{Omer Bobrowski \thanks{omer@ee.technion.ac.il}}
\affil{Viterbi Faculty of Electrical Engineering\\Technion - Israel Institute of Technology}

\maketitle

\begin{abstract}
In this article we propose a novel approach for comparing the persistent homology representations of two spaces (filtrations). Commonly used  methods are based on numerical summaries such as persistence diagrams and persistence landscapes, along with suitable metrics (e.g. Wasserstein). These
summaries are useful for computational purposes, but they are merely a marginal of the actual topological information that persistent homology can provide.
Instead, our approach compares between two topological
representations directly in the data space. We do so by defining a correspondence
relation between individual persistent cycles of two different spaces, and devising a
method for computing this correspondence. Our matching of cycles is based on both
the persistence intervals and the spatial placement of each feature.
We demonstrate our new framework in the context of topological inference, where we use statistical bootstrap methods in order to differentiate between real features
and noise in point cloud data.
\end{abstract}


\section{Introduction}

Topological Data Analysis (TDA) refers to a set of methods and tools that promote the use of mathematical topology in analyzing data and networks \cite{carlsson_topology_2009,ghrist_barcodes:_2008,wasserman_topological_2018}. The key idea is that topology can be used to 
study the shape of data in a rather qualitative way that is coordinate-free and robust to continuous deformations.
One of the main mathematical objects used in TDA is called \emph{homology} -- an algebraic-topological structure that can be used to characterize shapes by their connected components (referred to as $0$-cycles), holes ($1$-cycles), cavities ($2$-cycles), and higher-dimensional generalizations of these ($k$-cycles). This article focuses on homology, as well as its multi-scale version known as \emph{persistent homology} \cite{zomorodian_computing_2005,edelsbrunner_persistent_2008}. See Section \ref{sec:prelims} for more details.

In recent years, persistent homology has been incorporated into machine learning tasks in various fields, including image analysis \cite{lawson2019persistent,QAISER20191}, shape analysis \cite{skraba_persistence-based_2010,li_persistence-based_2014,crawford2020predicting}, time-series data analysis \cite{perea_sliding_2014,emrani_real_2014}, design and analysis of deep neural networks \cite{hofer_deep_2017,hu2019topology,pmlr-v108-gabrielsson20a,naitzat_topology_deep}, and more. 
The rise in its popularity can be attributed to two main factors. The first is its ability to produce a low-dimensional compact representation of high-dimensional data patterns. The second is the development of persistent-homology-based metrics, feature vectors and kernels, promoting the incorporation of persistent homology into machine learning models and tools. 
These developments include metrics such as  \emph{Wasserstein} and \emph{bottleneck} distances \cite{cohen-steiner_stability_2007,mileyko_probability_2011}, vector representations such as  \emph{persistence images} \cite{adams_pers_image_2017} and  \emph{persistence landscapes} \cite{bubenik_statistical_2007}, and kernels such as \emph{persistence scale space kernels}  \cite{kwitt_statistical_2015} and \emph{persistence weighted Gaussian kernels} \cite{kusano_persistence_2016}.
Commonly, persistent homology is used for extracting a topological ``signature'' from data as a pre-processing stage, prior to applying other machine learning or other statistical methods.
In addition, it can also be treated as an analysis tool, that can be used to assess the topological complexity of data when choosing a learning strategy \cite{bae_beyond_2017}.
For thorough review of persistent homology in machine learning and its applications see \cite{pun2018persistenthomologybased}.

\begin{figure}[h]
	\begin{subfigure}{0.5\textwidth}
	\centering
		\includegraphics[width=0.5\textwidth]{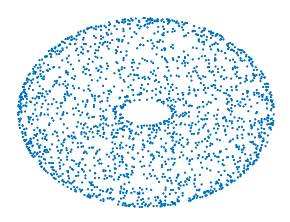}
		\caption{}
	\end{subfigure}
	~
	\begin{subfigure}{0.5\textwidth}
	\centering
		\includegraphics[width=0.8\textwidth]{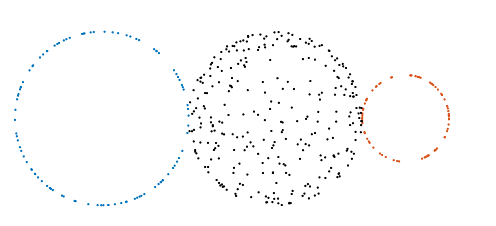}
		\caption{}
	\end{subfigure}
	~
	\begin{subfigure}{0.5\textwidth}
	    \centering
		\includegraphics[width=0.8\textwidth]{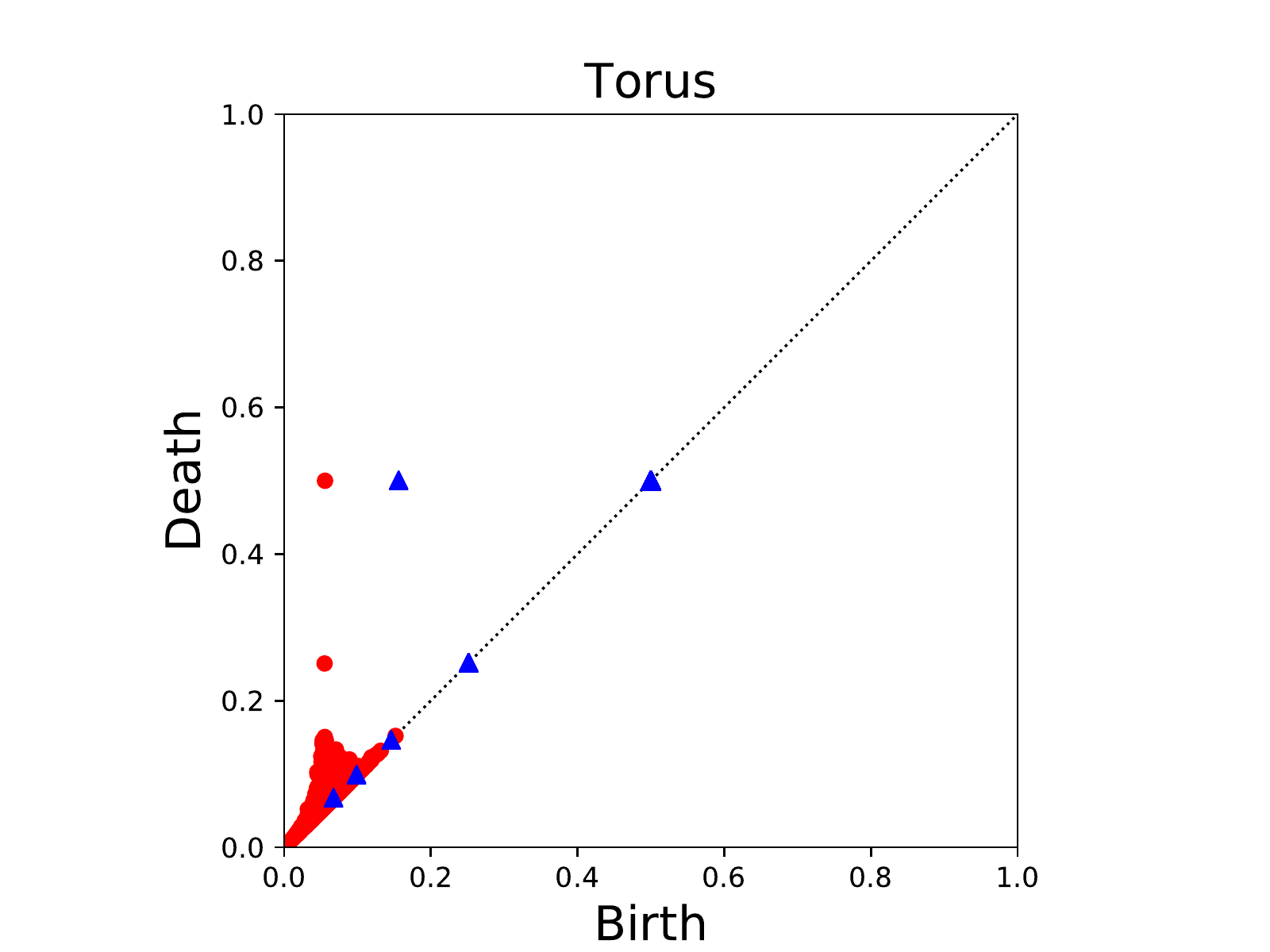}
		\caption{}
	\end{subfigure}
    ~
	\begin{subfigure}{0.5\textwidth}
	    \centering
		\includegraphics[width=0.8\textwidth]{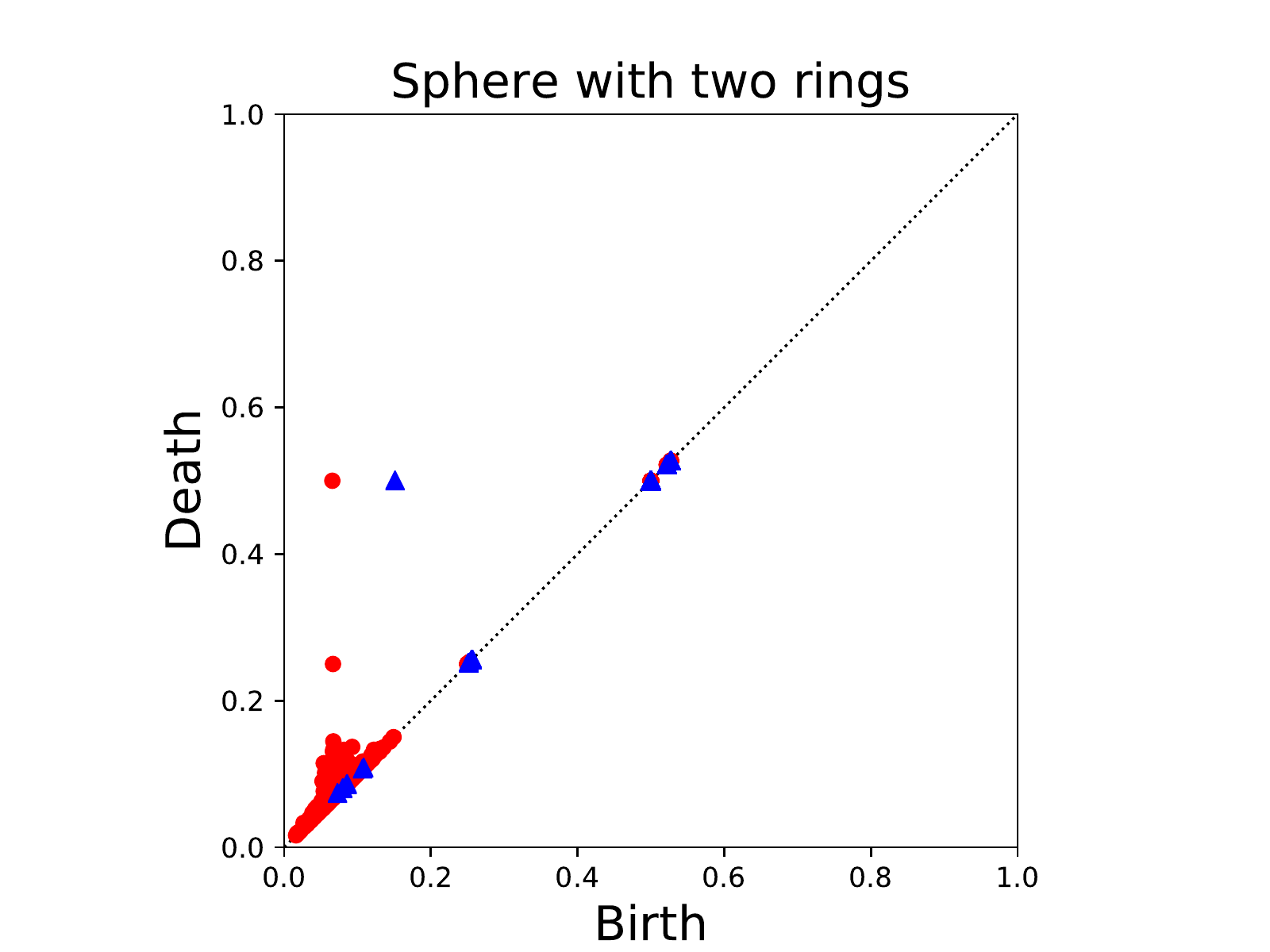}
		\caption{}
	\end{subfigure}
	\caption{(a) A random point cloud on a  torus. (b) A random point cloud sampled uniformly from a 2D sphere (black points) with two circles attached to it. (c)-(d) The persistence diagrams generated for these point clouds by taking the union of balls with an increasing radius. The red dots represent the first homology (holes) and the blue triangles represent the second homology (bubbles). While the point clouds are sampled from two different topological spaces, they produce similar persistence diagrams.}
	\label{fig:diff_spaces_same_hom}
\end{figure}

\begin{figure}[h]
\centering
	\begin{subfigure}{0.4\textwidth}
		\centering		
		\includegraphics[width=\textwidth]{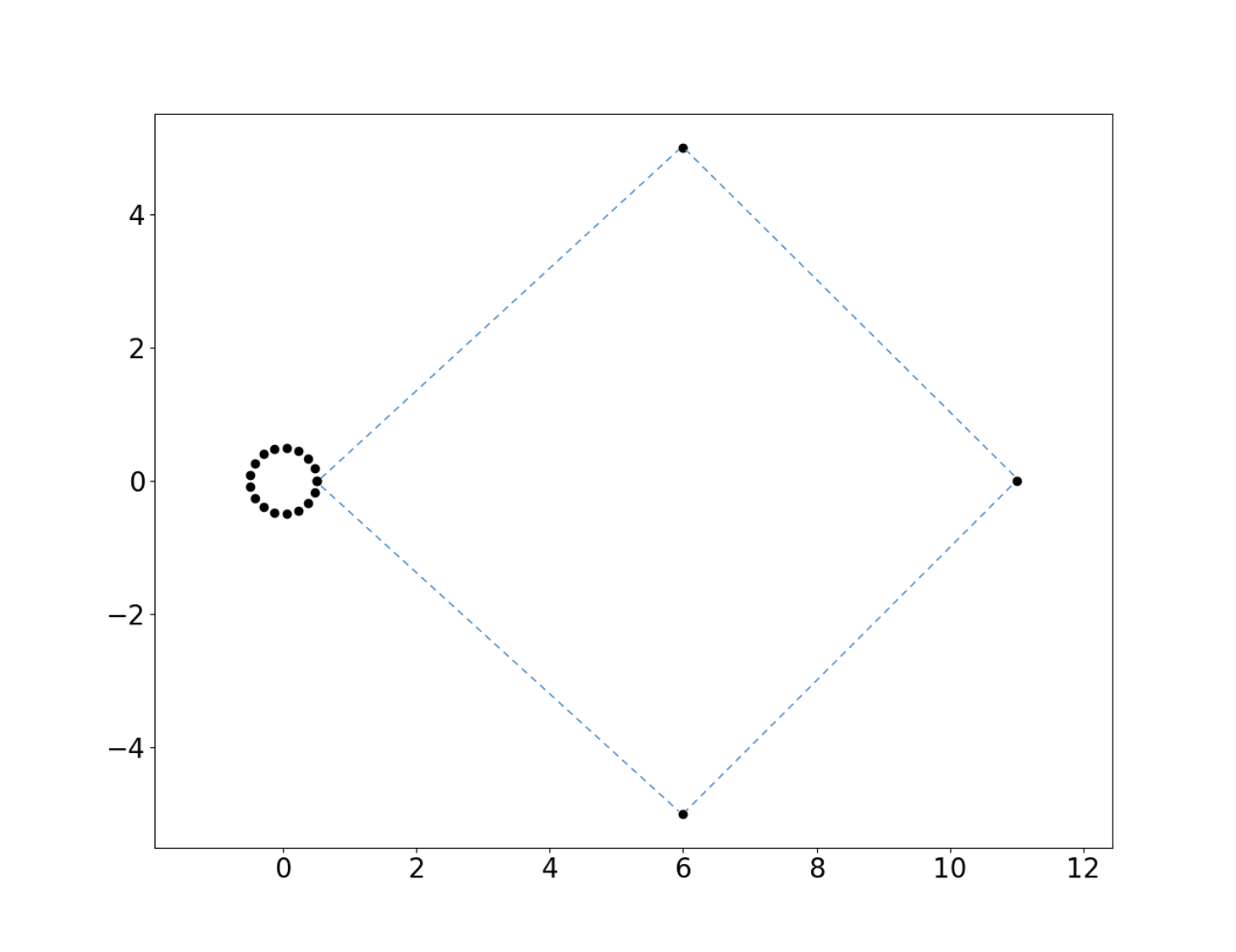}
		\caption{}
		\label{fig:bad_a}
	\end{subfigure}
	\begin{subfigure}{0.4\textwidth}
		\centering		
		\includegraphics[width=\textwidth]{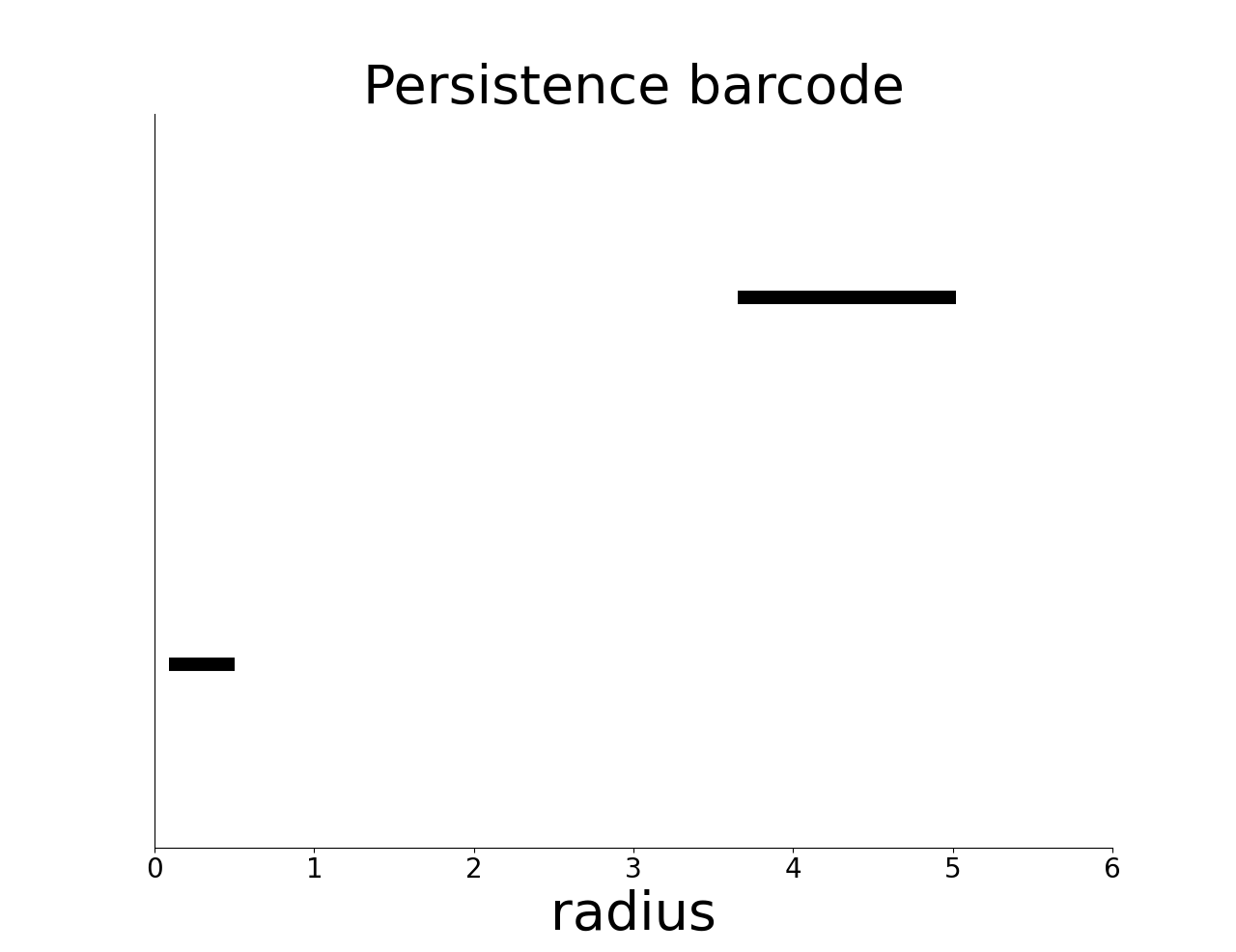}
	\caption{}	
	\end{subfigure}
\caption{(a) Circle with outliers. (b) Persistence barcode of the first homology (holes) generated by taking the union of balls with an increasing radius. Each bar corresponds to the lifespan of a $1$-cycle. While the cycle generated by the small dense circle is intuitively more significant,
the cycle generated by the three single points (highlighted by the dashed lines) has a longest lifespan. 
}
\label{fig:bad_example}
\end{figure}

In the course of inferring the homology of a given data set, a fundamental question that arises is --  
how can one differentiate between the ``signal'' (i.e.~meaningful features underlying the data) and ``noise''? In order to be able to provide such a distinction, a key requirement is to be able to compare the homology of different data sets. In the context of persistent homology, the most commonly used measures of comparison are the \emph{bottleneck} and the \emph{Wasserstein} distances. These distances (as well as others) have several advantages, including suitable notions of stability that can be proved and exploited \cite{cohen-steiner_stability_2007,cohen-steiner_lipschitz_2010,skraba2020wasserstein}. However, the use of these metrics in statistical applications has two key disadvantages.
The first one is that the input for these metrics is a numerical marginal of persistent homology (i.e.~\emph{persistence diagrams} or \emph{barcodes}), rather than the topological features themselves. Briefly, a persistence diagram summarizes  the information in persistent homology using a collection of $(birth, death)$ pairs representing the range of scales at which  each feature was observed. 
Comparing persistence diagrams is appealing due do their simple numerical nature. However, this simple nature also means that we reduce the intricate algebraic information in persistent homology to a set of numbers, stripped of their original topological meaning. As a consequence, it is possible that two data sets that are completely different topologically will produce similar persistence diagrams, with a low Wasserstein distance, for example (see Figure \ref{fig:diff_spaces_same_hom}). The second disadvantages of using these metrics is that in many cases it involves a considerable amount of heuristics that are not necessarily justified statistically. For example, in many applications, the ``significance'' of topological features is determined by the length of their lifetime. While such measures might work in some cases, they can completely fail in others (see example in Figure \ref{fig:bad_example}).
The main goal of this work is to provide a mathematical and algorithmic framework, that will enable 
the design of robust topological-statistical methods, that in particular will avoid these two significant drawbacks.

The framework we present in this paper is based on the following key idea. Suppose that we have two topological spaces $X$ and $Y$. In order to determine their ``topological similarity'', rather than using a numerical distance, we will search for the topological features both spaces share in common. We refer to this approach as \emph{cycle matching} or \emph{cycle registration}.
In order to perform the topological feature matching, we use a third object $Z$ that serves as a common frame of reference, and into which the features in both $X$ and $Y$ can be mapped. For example, we can take $Z=X\cup Y$, into which the features of $X$ and $Y$ can be mapped using the inclusion map. Two motivating examples where our framework could be used are the following.

\begin{figure}[t]
\centering
	\includegraphics[width=0.9\textwidth]{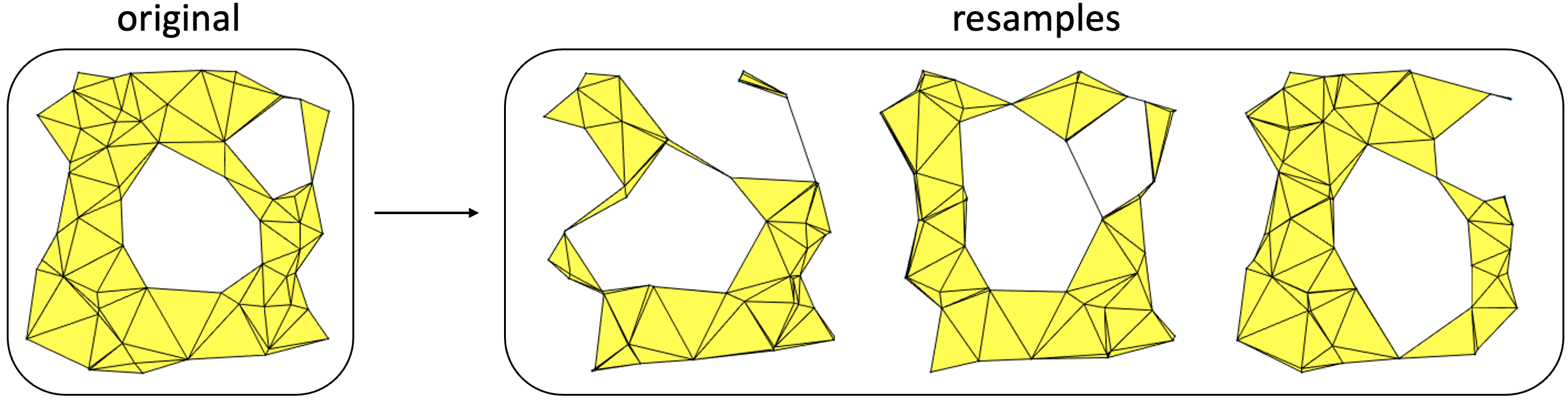}
    \caption{Bootstrap example. The original sample on the left contains two visible $1$-cycles. On the right we generated  three additional resamples. Visually one would argue that the large cycle in the middle reappears in all the new samples while the smaller cycle reappears only in the second. In this work we make this intuition into a formal mathematical framework.}
    \label{fig:boot_illus}
\end{figure}

\begin{figure}[t]
    \centering
    \includegraphics[width=0.7\textwidth]{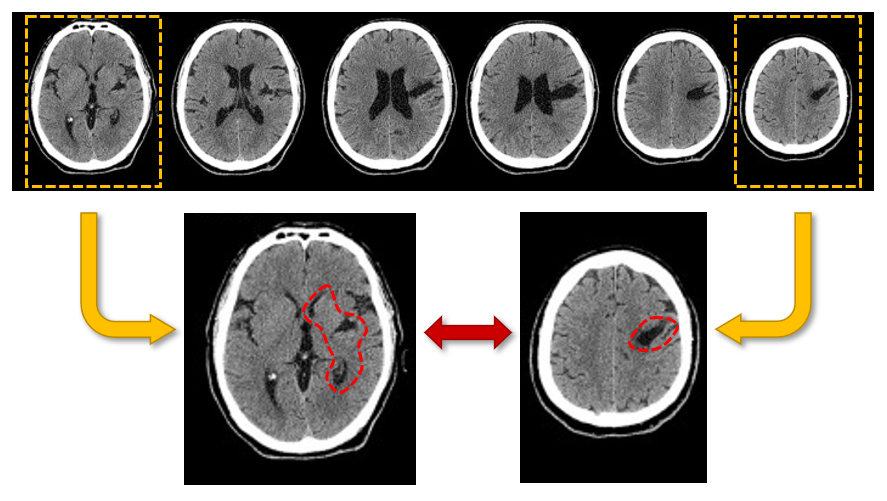}
    \caption{Top -- Consecutive slices of CT scans illustrating right frontal, temporal, and parietal lesions. Bottom -- The left and right slices have a common topological feature, emphasized by the dashed loops, which corresponds to the lesions (top image is taken from \cite{lanzoni_focal_2014}).}
    \label{fig:CT_example}
\end{figure}

\begin{enumerate}
\item {\bf Statistical bootstrap.} 
Following the question of differentiating signal from noise presented above, the goal here is to estimate the probability that an observed feature was created by chance (i.e.~not due to any meaningful phenomenon intrinsic to the system), by using bootstrap-type methods. In other words, we want to generate a pool of resamples from the original data, and calculate the empirical probability of a given feature to appear across the resamples, see Figure \ref{fig:boot_illus}. The key challenge here, is that we need to provide a way to determine whether an observed feature in a resample is ``the same feature'' as the one observed in the original sample. This example is in fact the main motivation for this work, and will be explored  in Section \ref{sec:hom_inf}.

\item {\bf Medical imaging.} Suppose that we are provided with  a sequence of images of 2D slices in the brain (via CT, for example, see Figure \ref{fig:CT_example}), and we identify suspicious features in some of them. While these features can appear in different areas of the image, have different shapes and scale, we want to determine whether they are all part of the same topological pathology. If we can embed all the 2D images into a common 3D coordinate system, then the methods we present here would identify the matching features between different slices. We note that this example is illustrative only, and while it is highly interesting, it will not be pursued in this article.
\end{enumerate}

The key idea underlying our framework is the following. Given spaces $X$ and $Y$ with mappings into $Z$, we say that a cycle in $X$ matches a cycle in $Y$ if they are both mapped to the same (non-trivial) cycle in $Z$. This idea can be viewed as a 1-step \emph{zigzag persistence} \cite{carlsson_zigzag_2010}.
When using homology, defining and implementing this idea is relatively simple (see Section \ref{sec:cyc_match}). However, when using \emph{persistent} homology, both the definition and the implementation become  significantly more challenging. The main reason is that cycles that intuitively should be matched might have different birth/death times. 
The way we overcome this difficulty is by using \emph{image-persistent homology} (\cite{cohen_steiner_pers_image_2009}) that provides the appropriate link between persistent cycles in $X$ to those in $Y$.

Our cycle-registration framework is phrased in terms of general  topological spaces and filtrations. However, 
in order to provide algorithms implementing this framework, some discretization is required. Thus, the algorithms we provide are phrased in terms of \emph{simplicial complexes}, and \emph{simplicial homology}, which are at the heart of most of the methods, tools and applications in TDA. Briefly, a simplicial complex can be thought of as a high-dimensional graph, that in addition to vertices and edges may also includes triangles, tetrahedra, and higher dimensional simplexes.
Finally, in order to demonstrate the power of our cycle-registration framework we present the example of \emph{geometric} complexes. These are simplicial complexes generated by point clouds, as an approximation to its underlying structure. 
In this case, matching cycles should intuitively describe phenomena with a similar geometry (see Figure \ref{fig:boot_illus}).
We will demonstrate the powerful potential of our framework by studying the motivating application for this work -- bootstrap methods for persistent homology of geometric complexes. We also present a new metric for persistence diagrams, similar in spirit to the Wasserstein distance, but such that features across diagrams can only be compared if they represent matching cycles. 

\paragraph{Related work.}
The study in this article was partly inspired by \cite{carlsson_zigzag_2010,tausz_applications_2011} where an inclusion-based method was proposed to perform a bootstrap-type estimation for the significance of cycles.
The idea suggested there was to take a sequence of resamples and use zigzag persistence in order to find meaningful cycles.
There are a couple of key differences between the approach in \cite{tausz_applications_2011} and the one presented here. Firstly, the zigzag-based framework will be able to reveal features that exist \emph{only} through consecutive sequences of resamples. However, if a cycle is occasionally missing in one of the resamples, the zigzag framework will fail to identify all the instances as the same feature. Secondly, the zigzag scheme works at the static  homology level, while our framework applies to persistent homology as well.

Another use of bootstrap methods to assess statistical significance in persistent homology was studied in \cite{chazal_robust_2017,chazal_subsampling_2015,fasy_confidence_2014}. In these papers the authors present a framework for estimating confidence sets for persistence diagrams using the bottleneck distance. The differences between this work and ours are twofold.
Firstly, they aim to estimate significance for points in persistence diagrams (i.e.~the numeric $(birth, death)$ values), while we aim to estimate significance for the actual cycles (i.e.~the algebraic object). In other words, the framework in \cite{fasy_confidence_2014} will not be able to distinguish between two different cycles with similar $(birth, death)$ parameters (see Figure \ref{fig:diff_spaces_same_hom}), while ours is designed to do exactly that.
Secondly, the confidence sets in \cite{fasy_confidence_2014} are calculated for the entire persistence diagrams as a whole, whereas our method assesses a measure of significance for individual cycles of interest separately.

A parametric approach for statistical topology was suggested in \cite{adler2017modeling}, where the authors provide a parametric model for persistence diagrams. The main idea there is that given the persistence diagram of a given data set, one can estimate the model parameters, and utilize it in order to generate additional persistence diagrams with similar statistical properties as the original one. Then, these diagrams can be used in order to apply hypothesis testing and calculate statistical measures such as p-values. This approach is very efficient computationally, as it saves the need to repeatedly calculate the persistent homology of new samples. This framework differs from the one proposed here in the same ways as the ones mentioned in the previous paragraph.

Finally, topological inference and the challenge of separating signal from noise was addressed in a more probabilistic approach in \cite{bobrowski_topological_2017,chazal_convergence_2015,niyogi_noisy_2011,pokorny_persistent_2012}. These works provide useful theoretical analyses and guarantees. However, they require several assumptions on the distribution generating the data which here we would like to avoid.

\paragraph{Outline.} 
In section \ref{sec:prelims} we give a brief introduction to the terms and  structures discussed in this paper. In Section \ref{sec:goals} we formally state the main goals and challenges. 
Section \ref{sec:registration}  presents the generic framework for cycle registration in homology and persistent homology.
In Section \ref{sec:alg} we focus on simplicial complexes, in order to provide algorithms for  computing the different elements in our cycle-matching scheme.
Finally, in Section \ref{sec:app} we demonstrate the use of our framework in two areas.  The first is in the context of homology inference, using the bootstrap-like method discussed above. The second is in proposing a new metric for persistence diagram, inspired by the Wasserstein distance.

\section{Preliminaries}
\label{sec:prelims}

In this section we give a brief introduction to simplicial complexes, homology and persistent homology. For more details, see for example \cite{edelsbrunner_topological_2002,edelsbrunner_persistent_2008,edelsbrunner_persistent_2014,hatcher_algebraic_2002,zomorodian_computing_2005}. 

\subsection{Simplicial homology}

An \emph{abstract simplicial complex} over a set $S$, is a collection of finite subsets $K$ that is closed under inclusion, i.e., if $P\in K$ and $Q\subset P$, then, $Q\in K$. The elements of $K$ are called \emph{simplexes} and their dimension is determined by their size minus one. For $Q\subset P\in K$, we say that $Q$ is a \emph{face} of $P$, and $P$ is a \emph{co-face} of $Q$ of \emph{co-dimension} $d$, where $d=\dim(P)-\dim(Q)$.  

\begin{figure}[h]
    \centering
    \includegraphics[scale=0.6]{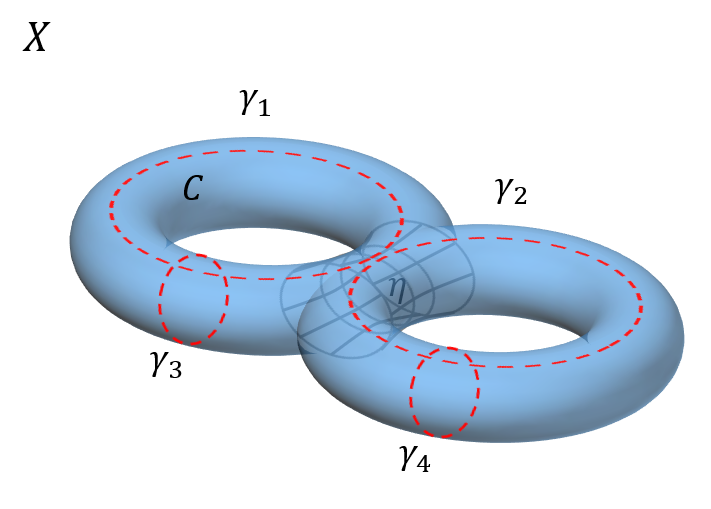}
    \caption{Homology example. The space here is 2-dimensional surface of genus 2. This manifold is composed of a single connected component, denoted by $C$, has four $1$-dimensional holes denoted by $\gamma_1, \gamma_2, \gamma_3$ and $\gamma_4$ (emphasized by the red dashed circles), and one $2$-dimensional hole, denoted by $\eta$, which is the entire surface that encloses the air-pocket inside. Hence, loosely speaking the homology groups of $X$ are given by $H_0(X)=\spn\{C\}$, $H_1(X)=\spn\{\gamma_1,\gamma_2,\gamma_3,\gamma_4\}$, $H_2(X)=\spn\{\eta\}$ and $H_k(X)=0$ for $k\geq3$.}
    \label{fig:Hom_example}
\end{figure}

\paragraph{Homology.}
Homology is a topological-algebraic structure that describes the shape of a topological space by its connected components, holes, cavities, and generally  $k$-dimensional cycles (see Figure \ref{fig:Hom_example}). 
Loosely speaking, given a topological space $X$, $H_0(X)$ is an abelian group generated by elements that correspond to the connected components of $X$; similarly $H_1(X)$ is generated by ``holes'' in $X$; $H_2(X)$ is generated by the ``cavities'' or ``bubbles'' in $X$. Generally, we can define the group $H_k(X)$ generated by the non-trivial $k$-dimensional cycles of $X$. A $k$-dimensional cycle can be thought of as the boundary of a $(k+1)$-dimensional object (i.e.~with the interior excluded).

Formally, let $X$ be a simplicial complex. The $k$-dimensional chain group $C_k(X)$ is a free abelian group generated by the $k$-dimensional simplexes in $X$. 
In this article we will use $\mathbb{Z}_2$ coefficients, and therefore $C_k(X)$ is a vector space. The elements of $C_k(X)$ are formal sums of $k$-simplexes called \emph{chains}.
The \emph{boundary homomorphism} $\partial_k:C_k(X)\to C_{k-1}(X)$ is defined as follows. If $\sigma$ is a chain representing a single $k$-simplex, then
$
	\partial_k(\sigma) = \sum_{\tau < \sigma}\tau$,
where $\tau < \sigma$ denotes that $\tau$ is a $(k-1)$-dimensional face of $\sigma$.
For general chains $\gamma=\sum_{i}\sigma_i\in C_k(X)$, $\partial_k$ extends linearly, i.e.~
$
    \partial_k(\gamma)=\sum_{i}\partial_k(\sigma_i)$.
It can be shown that $\partial_{k-1}\circ\partial_k \equiv 0$ for every $k>0$, and the sequence 
\[
\cdots\rightarrow C_{k+1}(X)\overset{\partial_{k+1}}{\rightarrow} C_k(X) \overset{\partial_k}{\rightarrow} C_{k-1}\rightarrow\cdots
\]
is known as a \emph{chain complex}.
Next, we define the subgroups
\[
Z_k(X)=\Ker(\partial_k(X)),\qquad
B_k(X)=\Imm(\partial_{k+1}(X)),
\]
so that $B_k(X)\subset Z_k(X)$. The group $Z_k(X)$ is known as the \emph{$k$-cycle group} (i.e.~chains whose boundary is zero) and $B_k(X)$ as the \emph{$k$-boundary group} (i.e.~$k$-cycles that are boundaries of $(k+1)$-dimensional chains).
The $k$-th \emph{homology group} is then defined as the quotient group,
$$H_k(X)= Z_k(X) / B_k(X).$$
In other words, the $k$-th homology group $H_k(X)$ consists equivalence classes of $k$-dimensional cycles who differ only by a boundary (called \emph{homological} cycles).
The ranks of the homology groups, called the \emph{Betti numbers}, are denoted
$\beta_k = \rank(H_k)$.

As mentioned earlier, intuitively speaking, the generators (or basis) of $H_0(X)$ correspond to the connected components of $X$, $H_1(X)$ corresponds to the holes in $X$, and $H_2(X)$ are the cavities. The definitions provided above are for simplicial homology, while other notions of homology groups can be defined for a much larger classes of topological spaces (see \cite{hatcher_algebraic_2002}). The intuition, however, is similar. 

In addition to homology, throughout the paper we will also use the following two terms which can be defined for simplicial homology as well as the more general notions of homology.

\paragraph{Induced homomorphisms.} Let $X$ and $Y$ be topological spaces and let $f:X\rightarrow Y$ be a continuous function.
Then homology theory provides a sequence of induced functions denoted $f_* :H_k(X)\rightarrow H_k(Y)$, that map $k$-cycles in $X$ to $k$-cycles in $Y$.

\paragraph{Homotopy Equivalence.} This is a notion of similarity between spaces that is weaker than homeomorphism.
Loosely speaking, two topological spaces $X$ and $Y$ are \emph{homotopy equivalent}, denoted $X\simeq Y$, if one can be continuously deformed into the other. 
In particular, homotopy equivalence between spaces implies similar homology, i.e.~if $X\simeq Y$, then $H_k(X)\cong H_k(Y)$ for all $k\geq 0$.

\subsection{Geometric complexes}
\label{sec:geo_comp}
Simplicial complexes are the fundamental building blocks in many TDA methods, where they are used for approximating geometric shapes using discrete structures. While the framework developed in this article can be applied to abstract topological spaces, we will focus on the case of simplicial complexes, and in particular geometric complexes, as these are the main motivation for our study.
In this section we present a few  special types of geometric  complexes commonly used in TDA.

\begin{definition}[\v{C}ech Complex]\label{def:cech}
Let $\cX$ be a finite set of points in a metric space. The \emph{\v{C}ech} complex of $\cX$ with radius $r$, denoted by $\Cech_r(\cX)$, is an abstract simplicial complex, constructed using the intersections of balls around $\cX$,
$$\Cech_r(\cX) :=  \big\{ Q\subset\cX : \bcap_{x\in Q} B_r(x)\neq\emptyset \big\},$$
where $B_r(x)$ is a ball of radius $r$ centered at $x$. 
\end{definition}

\begin{definition}[Vietoris-Rips Complex]\label{def:rips}
Let $\cX$ be a finite set of points in a metric space. The \emph{Vietoris-Rips} complex of $\cX$ with radius $r$, denoted by $\Rips_r(\cX)$, is an abstract simplicial complex, constructed  by pairwise intersections of balls,
$$\Rips_r(\cX) :=  \big\{Q \subset \cX : \forall x, x'\in Q\text{, }B_r(x)\cap B_r(x')\neq\emptyset\big\}.$$
In other words, all the points in $Q$ are within less than distance $2r$ from each other.
\end{definition}

\begin{figure}
     \centering
     \begin{subfigure}[b]{0.3\textwidth}
         \centering
         \includegraphics[width=\textwidth]{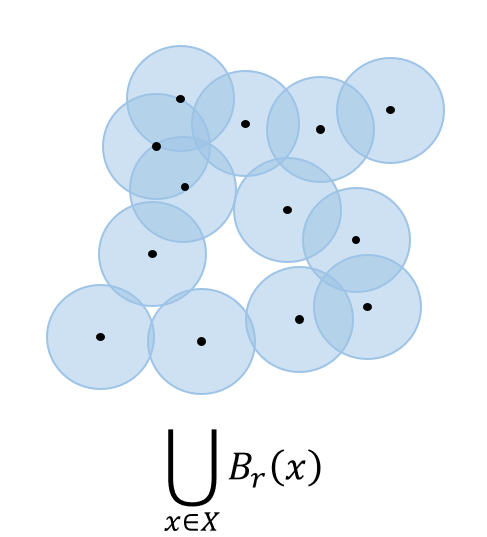}
        \caption{}
     \end{subfigure}
     \begin{subfigure}[b]{0.3\textwidth}
         \centering
         \includegraphics[width=\textwidth]{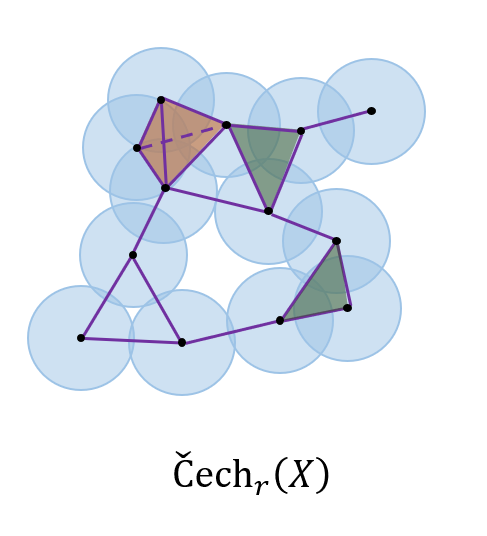}
         \caption{}
     \end{subfigure}
     \begin{subfigure}[b]{0.3\textwidth}
         \centering
         \includegraphics[width=\textwidth]{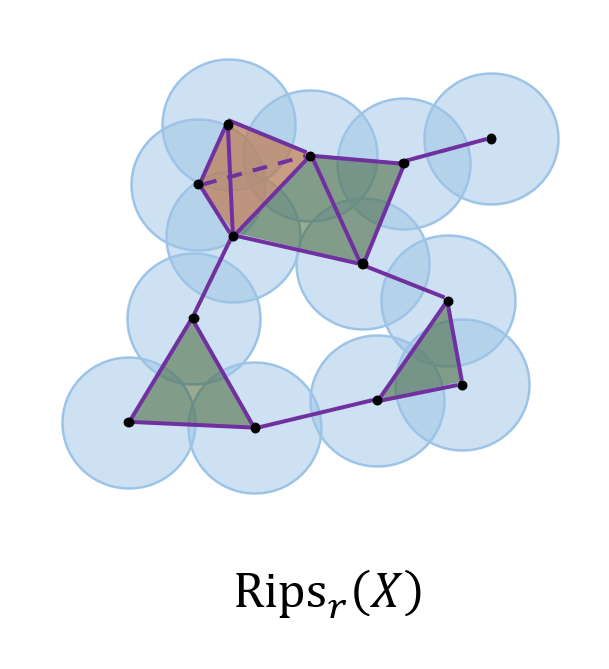}
         \caption{}
     \end{subfigure}
     \caption{(a) $\bigcup_{x\in X}B_r(x)$ - the ball cover of $X$ with radius $r$. (b) $\Cech_r(X)$ - the \v{C}ech complex of $X$ with radius $r$. (c) $\Rips_r(X)$ - the Vietoris-Rips complex of $X$ with radius $r$.}
        \label{fig:Cech_Rips}
\end{figure}

The \v{C}ech and Rips complexes are the most extensively studied complexes in TDA.
The Rips complex is commonly used in applications (see for example \cite{de_silva_homological_2007})  due to its simple definition that depends on pairwise distances only. The Rips complex can be also viewed as an approximation for the \v{C}ech complex by the following relation \cite{de_silva_coverage_2007,edelsbrunner_computational_2010},
$$\Rips_r(\cX)\hookrightarrow\Cech_{\sqrt{2}r}(\cX)\hookrightarrow\Rips_{\sqrt{2}r}(\cX).$$
The construction of the \v{C}ech complex is a bit more intricate, hence it is slightly less popular in applications. However, the \v{C}ech complex plays a central role in many theoretical results, especially in the random setting (see for example \cite{kahle2011random,yogeshwaran2017random,bobrowski2019homological,auffinger2020topologies,skraba2020homological}). This largely due to the fact that the \cech complex is homotopy equivalent to the ball cover inducing it (due to the Nerve Lemma \cite{borsuk_imbedding_1948}).

\subsection{Persistent homology}
Persistent homology is one of the fundamental tools used in TDA, and can be thought of as a multi-scale version of homology.
While homology is calculated for a single space $X$, persistent homology is applied to a \emph{filtration}.
Let $X$ be a topological space and consider a filtration $\{X_t\}_{t\in \mathbb{R}}$, so that for all $s\le t$ we have $X_s\subset X_t\subset X$. As $t$ is increased, holes can be created and/or filled in, demonstrating changes to the homology. Persistent homology is used to track these changes.

For $s\le t$, the inclusion $X_s \hookrightarrow X_t$ induces a homomorphism $H_k(X_s)\to H_k(X_t)$ between the homology groups. 
These induced maps enable us to track the evolution of homology classes throughout the filtration, from the point when they are first formed (\emph{born}) to the point when they become boundaries, and hence trivial (\emph{die}).
The algebraic structure tracking this evolution is called a \emph{persistence module}, denoted $\mathrm{PH}_k(X)$. In \cite{zomorodian_computing_2005} it was shown that $\mathrm{PH}_k(X)$ has a unique decomposition into basis elements called \emph{persistence intervals}. Intuitively, each persistence interval tracks a single $k$-cycle from birth to death.
For each persistence cycle $\gamma\in\PH_k(X)$ we denote by 
$\bth(\gamma)$ the point (value of $t$) where $\gamma$ is first created, and by $\dth(\gamma)$  the point where $\gamma$ becomes trivial. The entire lifetime interval is denoted by $\Int(\gamma)=[\bth(\gamma),\dth(\gamma))$. 
\begin{figure}
    \centering
    \includegraphics[scale=0.7]{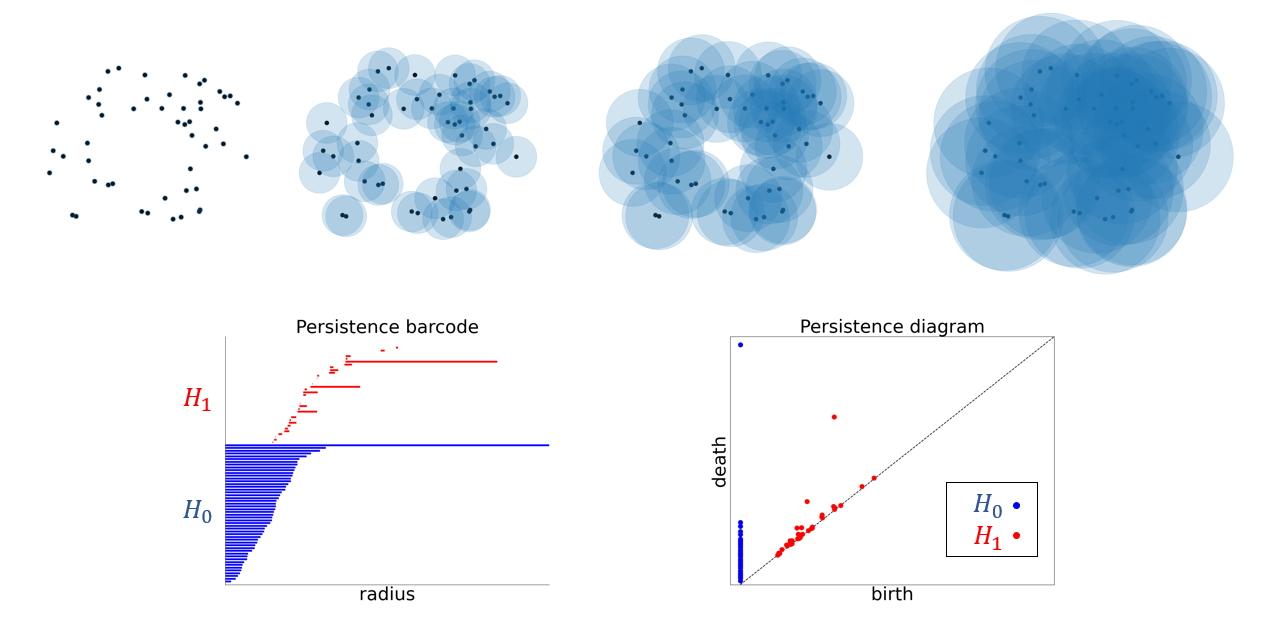}
    \caption{Persistent homology -- example. Top: a continuous filtration induced by inflating the ball cover around a point cloud. Bottom left: The resulting barcode summary for $\PH_0$ and $\PH_1$. Bottom right: The resulting persistence diagram. This example highlights the intuition that long bars (or points far from diagonal) stand for the topological features of the space underlying the point cloud. }
    \label{fig:Pers_hom_exmp}
\end{figure}

In most TDA applications, once persistent homology is calculated one outputs a numerical summary in the form of a \emph{barcode} or a \emph{persistence diagram} (see Figure \ref{fig:Pers_hom_exmp}). These are two equivalent ways to visually represent the collection of $(birth, death)$ pairs for all persistence intervals.
Note that while $\mathrm{PH}_k(X)$ is a rich algebraic structure,
the information provided by the barcodes (and persistence diagrams) strips it of all the algebraic information, and keeps only the marginal numerical values. Thus, significant information about the original spaces is lost. We will come back to this point later on.

An example where persistent homology is used is in the context of geometric complexes. Here, the filtration parameter is the radius $r$, and the persistent homology provides a summary for all the cycles that appear at different scales. Given point cloud data, we can compute the persistent homology of either the \cech or the Rips filtration in order to extract information about the topological space underlying the data.
The common approach  is to identify short bars in the barcode as noise and long bars (i.e.~cycles that persist over a long interval) as ``real'' or ``significant'' features (See Figure \ref{fig:Pers_hom_exmp}).

\subsection{Persistent homology computation}\label{sec:pers_hom_comp}
In section \ref{sec:cyc_match} we provide algorithms for intervals matching. In order to understand the way these algorithms operate, one needs to know the way persistent homology is calculated.   
For completeness, we briefly describe the classic algorithm for computing persistent homology. For more information, see for example \cite{zomorodian_topological_2012}.

Consider a finite filtration of abstract simplicial complexes $\mathcal{F}_X =\{X_m\}_{m=0}^M$.
In order to compute the persistent homology of $\mathcal{F}_X$, we represent the boundary operators $\{\partial_{k+1}\}_{k\geq 0}$ in a  matrix form, such that the rows and columns of $\partial_{k+1}$ are indexed by the $k$-simplexes, and $(k+1)$-simplexes, respectively.
We assume the rows and columns are sorted according to the filtration value of the simplexes -- the smallest index $m\in\{0,\ldots,M\}$ for which a simplex is in $X_m$ (if filtration values are identical, we choose the order arbitrarily). 
For simplicity, we consider homology groups with coefficients in $\mathbb{Z}_2$.
The entries of the matrix $\partial_{k+1}$ are then 
$$(\partial_{k+1})_{i,j} = 
	\begin{cases}
    		1       & \quad \text{the } i\text{-th } k \text{-simplex is a face of the } j \text{-th } (k+1)\text{-simplex},\\
    		0       & \quad \text{otherwise}.\\
  \end{cases}
$$
In addition, we add the following notation,
$$\low(j)=\min\{i:\forall i'>i,\  (\partial_{k+1})_{i',j}=0\}.$$
Since a column is associated with a $(k+1)$-simplex, the `low' value is the index of its face that is added last in the filtration.

\begin{figure}[h]
    \centering
    \includegraphics[width=\textwidth]{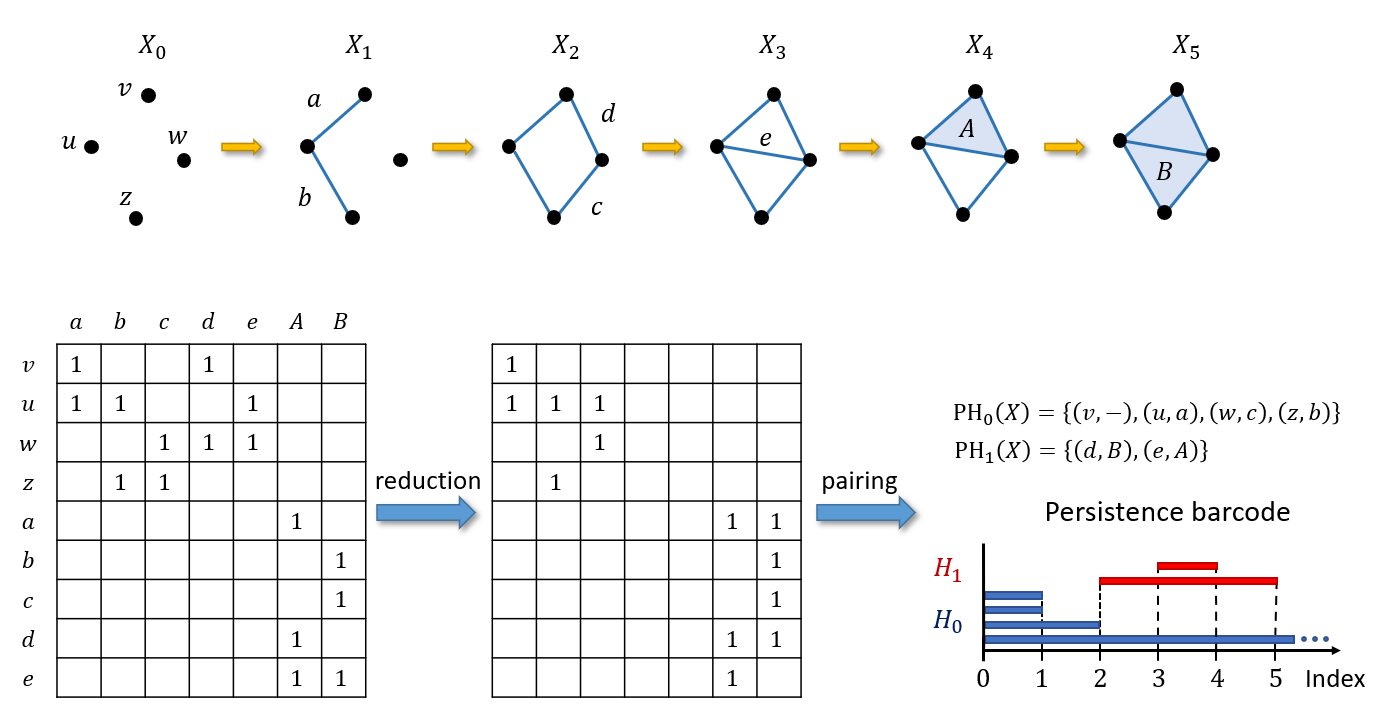}
    \caption{Persistent homology computation example. Top: $\mathcal{F}_X=\{X_m\}_{m=0}^{5}$ - a filtration of six spaces. Bottom: Persistent homology computation. We start with an extended boundary matrix ($\partial_0(X)$ and $\partial_1(X)$ joint together). The matrix is reduced using column operations. The result are pairs of (positive, negative) simplexes which relate to persistent intervals. In the example above, $\PH_0(X)$ (blue bars)  contains $3$ finite intervals that are generated by $u,w,z$ and eliminated by $a,c,b$ respectively, and one infinite interval that is generated by $v$. $\PH_1(X)$ (red bars) contains $2$ finite intervals that are generated by the chains $a+b+c+d$, $a+d+e$ and are eliminated by the chains $A+B$, $A$ respectively.}
    \label{fig:Pers_hom_comp}
\end{figure}

To compute persistent homology, we start by reducing the matrices $\{\partial_{k+1}\}_{k\geq 0}$ using column operations according to the following rules.
The $j$-th column can be added only to columns $j'$ with $j'>j$ and $\low(j)=\low(j')>0$.
The reduced boundary matrix $\tilde{\partial}_{k+1}$ can then be interpreted as follows.
For every column $j$ such that $\low(j)=0$,  the corresponding simplex  $\sigma_j$ is called \emph{positive}, which means that it can (depending on the filtration value) create a new nontrivial $(k+1)$-cycle. 
If $\low(j)>0$, then $\sigma_j$ is called \emph{negative}, which means that it potentially can mark the destruction of a $k$-cycle.
Finally, each negative $(k+1)$-simplex $\sigma_j$ is paired with the positive $k$-simplex $\tau_{i}$, that satisfies $i=\low(j)$. 
If $i<j$, then the interval $(i,j)$ represents a finite $k$-persistence interval in the persistent homology of $\mathcal{F}_X$, i.e.~a nontrivial $k$-cycle that is born in $X_i$ and dies in $X_j$. If $i=j$ we ignore the corresponding simplexes. 
A positive simplex in $\tilde{\partial}_{k}$ that does not have a matching negative simplex in $\tilde{\partial}_{k+1}$ generates an infinite $k$-persistence interval.
An example for the computation of persistent homology for a simple filtration can be seen in Figure \ref{fig:Pers_hom_comp}.

\section{Main Goals and Challenges}\label{sec:goals}

In many (possibly most) TDA applications, persistent barcodes are filtered based on  bar-length. While this is an intuitive action to take, the bar-length is mainly a heuristic measure, and using it in many cases is not  justified statistically. 
For example, consider the case in Figure \ref{fig:bad_example}.
In this paper we propose a different approach to analyzing the output of persistent homology, that relies on statistical properties of the features, rather than their size only. The key idea is the following.

Suppose that we take the point configuration in Figure \ref{fig:bad_example}, randomly resample it,
and compute the persistent homology. We repeat this process several times. While the cycle on the right in Figure \ref{fig:bad_a} has a large persistence value (long bar), it will rarely appear in the resamples.
On the other hand, the  cycle on the left will appear (in some version) in many of the resamples, even though its persistence value is small. Our approach is designed to capture such fundamental discrepancies.
We propose that such resample frequency measure of cycles, combined with the actual values from persistence diagrams, could provide a robust measure for assessing the importance of homological features in data.

While the framework we just described  sounds relatively simple, it is in fact quite challenging to implement. The main reason is the following. Given the original sample and a resample, in order to determine which cycles appear in both, one needs to provide a method to match cycles between two spaces. The problem is exacerbated, if one allows the resamples to also contain random perturbations.

The main goal of this paper is to provide a general framework for matching cycles (in both homology and persistent homology) between two spaces. We describe this framework for a general setting (i.e.~not specifically for geometric complexes) in Section \ref{sec:registration}. While the main motivation here is the resampling procedure described above, we believe that this framework can become useful in other applications as well, as stated in the introduction.

\section{Cycle Registration}
\label{sec:registration}

In this section we provide the formal definitions required to describe our cycle-registration scheme in its general form.
We first discuss the key idea for cycle-matching in fixed and persistent homology.
Then, we will present the main contribution of this article -- an \emph{interval matching} method for persistent homology.

\subsection{Cycle matching}\label{sec:cyc_match}

\begin{figure}
	\centering
	\includegraphics[width=0.7\textwidth]{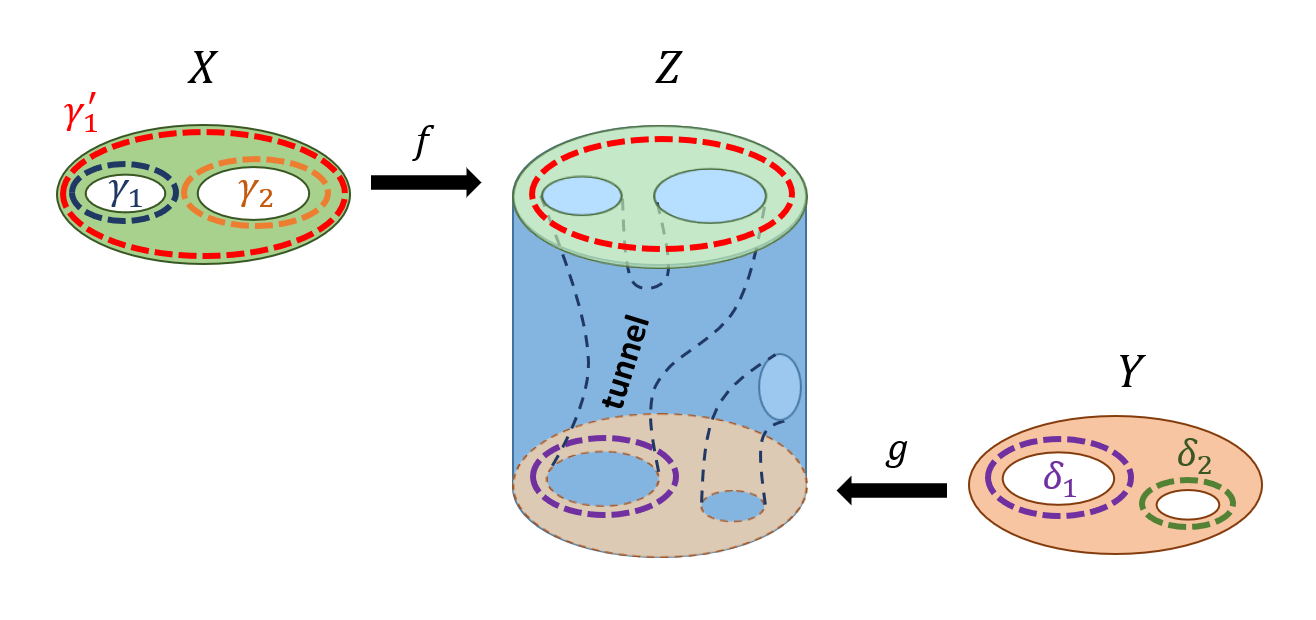}
	\caption{Cycle registration in fixed homology. The spaces $X$ and $Y$ are mapped to the top and bottom covers of the cylindrical space $Z$. The cycles $\gamma'_1$ and $\delta_1$ are matching cycles, since they are mapped to the same non-trivial cycle in $Z$. However, this matching scheme is sensitive to the basis choice. Choosing the basis $\gamma_1,\gamma_2$ for $H_1(X)$ results in no matches.}
        \label{fig:fixed_match}
\end{figure}

Let $X,Y$ and $Z$ be topological spaces.
Let $f:X\to Z$, and $g:Y\to Z$ be continuous functions, and denote by $f_*$ and $g_*$ the induced maps in homology, so that
$$H_k(X)\xrightarrow{f_*} H_k(Z) \xleftarrow{g_*} H_k(Y).$$
This gives rise to the following definition.
\begin{definition}
\label{defn:match_fixed}
Let $\gamma\in H_k(X)$ and $\delta\in H_k(Y)$. 
We say that $\gamma$ and $\delta$ match via $Z$ if 
$$[f_*(\gamma)]=[g_*(\delta)]\neq [0].$$
We denote this by $\gamma\overset{Z}{\sim}\delta$.
\end{definition}
In other words, we match cycles between $X$ and $Y$, by using a third space $Z$ as a frame of reference. See Figure \ref{fig:fixed_match} for a visual example.
Two useful cases for applying Definition \ref{defn:match_fixed} are:
 (a) taking $f,g$ as the inclusion maps (i.e.~when $X\cup Y\subset Z$), and (b) taking $f,g$ as projection functions (i.e.~when $Z \subset X\cap Y$). 

Consider the motivational example given in Figure \ref{fig:fixed_match}. As can be seen, different choices of a basis for $H_1(X)$ give different matches or no matches at all. 
Note that the straightforward approach to find all the matching $k$-cycles between $X$ and $Y$, is to take all possible combinations of basis elements of $H_k(X)$ and search all combinations of basis elements of $H_k(Y)$ for a possible match. 
In order to avoid such an exhaustive search, we propose a different approach. Instead of going from cycles in $X,Y$ to cycles in $Z$, we go in the opposite direction. From all the non-trivial $k$-cycles in $Z$ we select only those that are represented by both a generator in $f(X)$ and a generator in $g(Y)$. In other words, we find cycles in $Z$ that serve as ``tunnels'' from $f(X)$ to $g(Y)$ (with no opening in between), and we identify the openings of each such tunnel as matching cycles (see Figure \ref{fig:fixed_match}). 
By Definition \ref{defn:match_fixed}, these tunnels determine all possible matching cycles between $H_k(X)$ and $H_k(Y)$. 

Next, we want to extend Definition \ref{defn:match_fixed}, to perform a matching between cycles in \emph{persistent homology}.
In order to do so, we need to assume that the spaces $X,Y,$ and $Z$ are equipped with filtrations ${\cal{F}}_X = \{X_t\}_{t\in \R}$, ${\cal{F}}_Y = \{Y_t\}_{t\in \R}$, and ${\cal{F}}_Z = \{Z_t\}_{t\in \R}$.  In addition, we assume that there are functions $f_t:X_t\to Z_t$ and $g_t:Y_t\to Z_t$, such that $f_t|_{X_{s}}=f_s$ (the restriction of $f_t$ to $X_s$) and $g_t|_{Y_{s}}=g_s$. 
Hence, the following diagram commutes 
\begin{equation}\label{eqn:comm_diag}
\centering
\begin{tikzcd}
X_s \arrow[r,"i"] \arrow[d,"f_{s}"] 
	& X_t \arrow[d,"f_{t}"]\\
Z_s \arrow[r,"i"] 
	& Z_t
\end{tikzcd}
\end{equation}
Let $f_{t,*}$ and $g_{t,*}$ be the corresponding induced maps in homology.

\begin{definition}
\label{defn:match_pers}
Let $\gamma\in \PH_k({X})$ and $\delta\in \PH_k({Y})$. 
We say that $\gamma$ and $\delta$ match via ${Z}$, if there exists $t\in\Int(\gamma)\cap \Int(\delta)$ such that
$$[f_{t,*}(\gamma)] = [g_{t,*}(\delta)] \ne [0].$$
where $\Int(\cdot)$ denotes lifetime interval. We denote this by $\gamma\overset{\mathcal{F}_Z}{\sim}\delta$.
\end{definition}

Finding matching persistent cycles between $\PH_k(X)$ and $\PH_k(Y)$ can be done similarly to the fixed homology case. From all the persistent cycles of $\PH_k(Z)$ we select only those that for some fixed $t$ can be expressed by both a generator in $f_t(X_t)$ and a generator in $g_t(Y_t)$ (tunnel-like cycles, see Figure \ref{fig:fixed_match}). 

\begin{figure}[h]
    \centering
    \includegraphics[width=\textwidth]{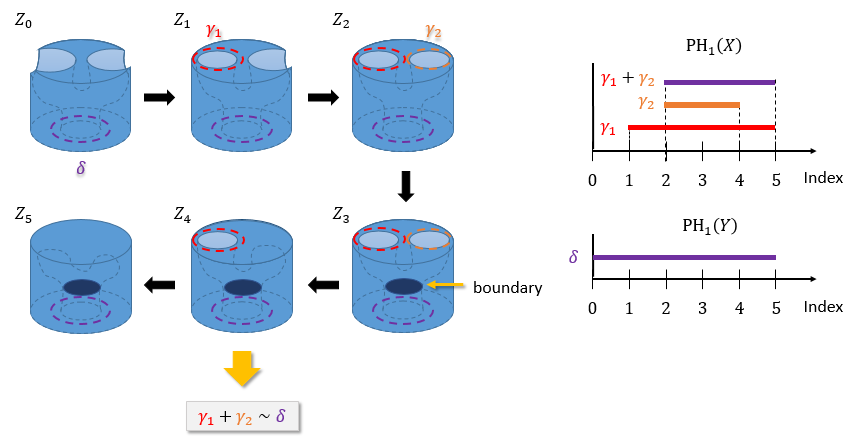}
    \caption{Persistent cycles vs.~persistence intervals. Left: A filtration $\mathcal{F}_Z=\{Z_i\}_{i=0}^5$. $X$ is the filtration induced by taking the top covers of the cylinders and $Y$ is induced by taking the bottom covers of the cylinders. Right: $\PH_1(X)$ is generated by $\gamma_1$ and $\gamma_2$, and $\PH_1(Y)$ is generated by $\delta$. Using Definition \ref{defn:match_pers} the only match between $X$ and $Y$ is $\gamma_1+\gamma_2\sim\delta$. However, $\gamma_1+\gamma_2$ is not one of the persistence intervals of $\PH_1(X)$. Note that by Definition \ref{defn:match_pd} we have $\gamma_2\overset{\mathcal{F}_{Z,\Int}}{\sim}\delta$.}
    \label{fig:pers_cycle_prob}
\end{figure}

Recall, that the standard basis elements of persistent homology $\PH_k$ are the \emph{persistence intervals}. Loosely speaking, these are the elementary persistent cycles whose birth and death values  generate  the persistence diagrams, and therefore are the most important in terms of applications. While Definition \ref{defn:match_pers} provides a matching between persistent \emph{cycles}, it is possible that it will fail to match  some of the persistence \emph{intervals} in $X$ and $Y$, as can be seen in Figure \ref{fig:pers_cycle_prob}. 
In the following, we want to develop a matching scheme  specifically for persistence intervals, that will allow us to address this issue.

\subsection{Persistence-intervals matching}\label{sec:ph_match}

We start by presenting the notion of \emph{image-persistent homology} presented in \cite{cohen_steiner_pers_image_2009} which serves a key role in our intervals matching framework.

To gain some intuition, we first consider the case of  fixed homology.
Let $X,Z$ be topological spaces, $f:X\to Z$ a continuous function, and let $f_*:H_k(X)\to H_k(Z)$ be the induced map. 
In this case, the image $\Imm(f_*)$ consists of cycles generated by chains in $X$ that are not mapped to boundaries in $Z$.
For example, take $Z=D^1$ to be the closed unit disk, $X=\partial D^1=S^1$ its boundary together with the natural inclusion map $f:X\hookrightarrow Z$, then, $H_1(X)\cong\mathbb{Z}_2$, but $\Imm(f_*)=0$, since the cycle in $X$ is mapped to a boundary in $Z$.

\begin{figure}
    \centering
    \includegraphics[width=\textwidth]{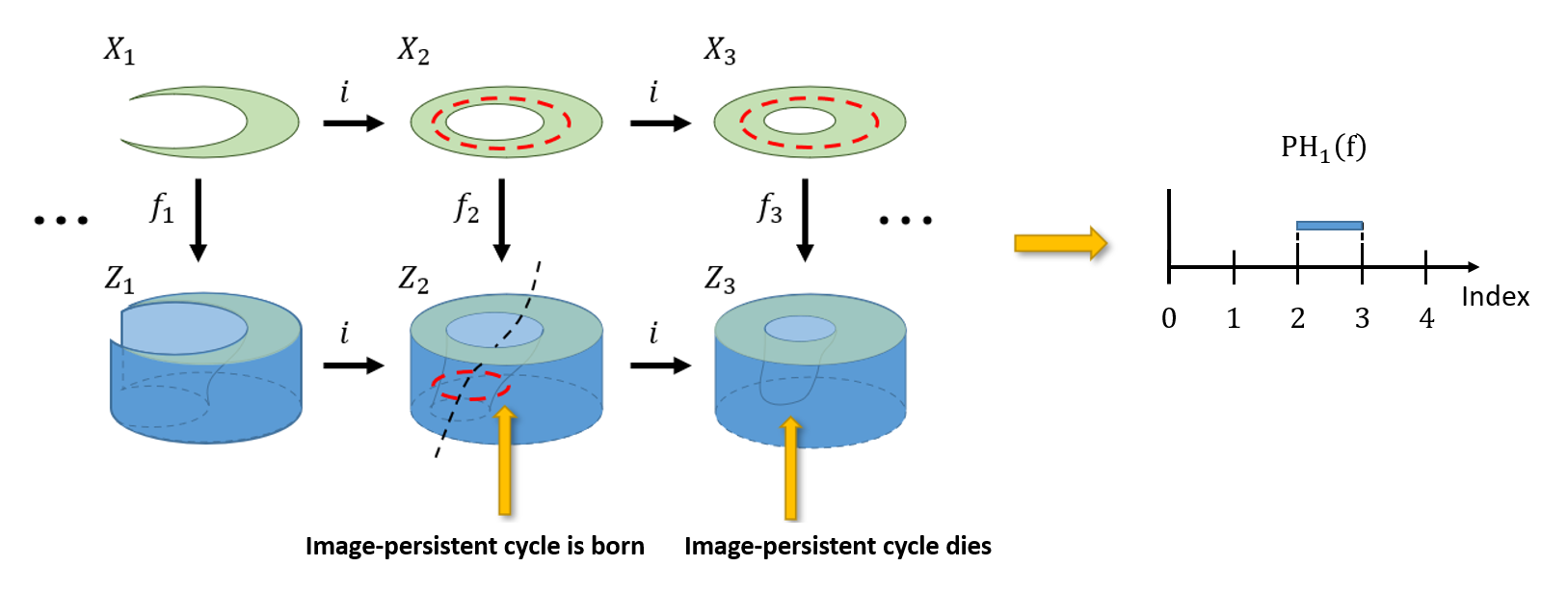}
    \caption{Image-persistence example. $X$ is a filtration induced by taking the top cylinder covers of the filtration of $Z$. An image-persistent cycle is born at time $2$, since a non-trivial cycle in $H_1(X_2)$ mapped to a non-trivial cycle in $H_1(Z_2)$. The cycle dies at time $3$ since it mapped to a boundary in $H_1(Z_3)$.}
    \label{fig:pers_proj_ex}
\end{figure}

Next, suppose that  $X$ and $Z$ are equipped  with the filtrations $\{X_t\}_{t\in\mathbb{R}}$ and $\{Z_t\}_{t\in\mathbb{R}}$, respectively.
In addition, suppose that for every $t\in\mathbb{R}$ we have a map $f_t:X_t\to Z_t$ such that $f_t|_{X_{s}}=f_s$ for $s\leq t$.  
For $s\leq t$, we have $X_s \subset X_t$, implying that the following diagram commutes
\begin{equation}\label{eqn:comm_diag_2}
\centering
\begin{tikzcd}
H_k(X_s) \arrow[r,"i_*"] \arrow[d,"f_{s,*}"] 
	& H_k(X_t) \arrow[d,"f_{t,*}"]\\
H_k(Z_s) \arrow[r,"i_*"] 
	&H_k( Z_t)
\end{tikzcd}
\end{equation}
Since $\Imm(f_{t,*})$ is a subgroup of $H_k(Z_t)$, and since the diagram commutes, one can use the restrictions of the maps $i_*:H_k(Z_s)\to H_k(Z_t)$ to define a persistence module on $\{\Imm(f_{t,*})\}_{t\in\mathbb{R}}$. This module is the image-persistent homology, denoted by $\PH_k(f)$. 
An example for image-persistent homology is given in Figure \ref{fig:pers_proj_ex}.
Next, we will utilize image-persistent homology in order to define a matching between persistence intervals.

\begin{remark}
In \cite{cohen_steiner_pers_image_2009} the authors focus on the case where $f_t = j_t:X_t \hookrightarrow Z_t$ are inclusion maps for all $t\in\mathbb{R}$, and provide an algorithm for calculating image-persistent homology in that case. However, as shown in the following, their framework is also applicable for any sequence of injective maps $\{f_t\}_{t\in\mathbb{R}}$ of the form defined above as well. 
To see that, we can separate $f_t$ into  two steps $f_t=f_t^2\circ f_t^1$ such that,
$$X_t\overset{f_t^1}{\rightarrow}\Imm f_t\quad\text{and}\quad\Imm f_t\overset{f_t^2}{\rightarrow}Z_t,$$
where $f_t^1=f_t:X_t\to{\Imm f_t}$, and $f_t^2:\Imm f_t\hookrightarrow Z_t$ is the inclusion  map. The second step is exactly the case discussed in \cite{cohen_steiner_pers_image_2009}.
Since $f_t$ is injective, the map $f_t^1$ in the first step is a bijection. Hence, the persistence module of $\{\Imm f_{t,*}\}$ is isomorphic to the persistence module of $\{\Imm f_{t,*}^2\}$ and can be computed from it.
\end{remark}

Before presenting the formal definition for matching intervals, we wish to explain the intuition behind it in the simplicial setting.
Given the setting in Definition $\ref{defn:match_pers}$, recall that from all the persistent cycles in $Z$ we are interested only in cycles that for some $t$, can be represented by both a generator in $f_t(X_t)$ and a generator in $g_t(Y_t)$. 
In order to identify such cycles, we compute the image-persistent homology for both $f$ and $g$.
Recall (see Section \ref{sec:pers_hom_comp}) that finite persistence intervals are associated with an eliminating negative simplex. In addition, the negative simplexes in both $\PH_k(f)$ and $\PH_k(g)$ are a subset of the negative simplexes in $\PH_k(Z)$. We utilize this fact in order to match finite intervals between $\PH_k(f)$ and $\PH_k(g)$, by checking if they are eliminated by the same negative simplex. Note that this does not apply for infinite intervals who do not have an eliminating simplex. We will show how to handle those at the end of this section.

Throughout, we will assume that the filtrations we are dealing with are Morse filtrations, as defined in the following.

\begin{definition}[Morse filtration]\label{defn:filt}
We say that a filtration $\{{X}_t\}_{t\in\R}$ is a \textbf{Morse filtration} if there exists a finite set $T_c\subset \R$, that satisfies the following.
\begin{enumerate}  
\item For all $t\not\in T_c$, there exists $\epsilon>0$ small enough so that for any $0<\epsilon'<\epsilon$, the map $i_*:H_k(X_{t-\epsilon'})\to H_k(X_{t+\epsilon'})$ induced by inclusion is an isomorphism. In other words, homology does not change at $t$.
\item For all $t \in T_c$, there exists $\epsilon>0$ small enough so that for any $0<\epsilon'<\epsilon$, either
\[
\begin{split}
	&i_*:H_k(X_{t-\epsilon'}) \to H_k(X_{t+\epsilon'}) \text{ is injective, and } \beta_k(X_{t+\epsilon'}) = \beta_k(X_{t-\epsilon'})+1, \text{ or}\\
	&i_*:H_k(X_{t-\epsilon'}) \to H_k(X_{t+\epsilon'}) \text{ is surjective, and } \beta_k(X_{t+\epsilon'}) = \beta_k(X_{t-\epsilon'})-1.
\end{split}
\]
In other words, the homology changes allowed are either the creation of a single new cycle, or the termination of a single existing cycle.
\end{enumerate}
\end{definition}

Let ${X, Y, Z}$ be topological spaces, equipped with filtrations $\mathcal{F}_X, \mathcal{F}_Y, \mathcal{F}_Z$, respectively.
Assume that the mappings $f_t,g_t$ satisfying \eqref{eqn:comm_diag}. The following definition provides our notion of matching between persistence intervals.

\begin{definition}
\label{defn:match_pd}
Let $\gamma\in\PH_k(X)$ and $\delta\in\PH_k(Y)$ be persistence intervals.
Suppose that there exist $\tilde{\gamma}\in\PH_k(f)$ and $\tilde{\delta}\in\PH_k(g)$ such that  
\[
\begin{split}
\bth(\gamma)&=\bth(\tilde{\gamma}),\\
\bth(\delta)&=\bth(\tilde{\delta}), \\
\dth(\tilde{\gamma})&=\dth(\tilde{\delta}).
\end{split}
\]
Then we say that $\gamma$ and $\delta$ are matching intervals via $Z$, denoted  $\gamma\overset{\mathcal{F}_{Z,\Int}}{\sim}\delta$.
\end{definition}

The third condition in Definition \ref{defn:match_pd} together with our assumption on Morse filtrations, imply that the intervals 
$\tilde{\gamma}$ and $\tilde{\delta}$ are eliminated by the same boundary element in $Z$.
The first two conditions imply that the intervals $\tilde \gamma, \tilde\delta$ are structurally related to $\gamma,\delta$, since they appear at the same filtration value.
For instance, in the simplicial case $\gamma$ and $\tilde{\gamma}$ are  generated by a positive simplex in $X$ and its image in $\Imm f$ respectively.
Notice that if $\gamma\overset{\mathcal{F}_Z}{\sim}\delta$ according to Definition \ref{defn:match_pers} then they also match according to Definition \ref{defn:match_pd}. The converse, however, is not true generally, as can be seen in Figure \ref{fig:pers_cycle_prob}. Note that the image-persistent cycles of $\PH_1(f)$ in this figure are generated by $\gamma_1$ and $\gamma_1+\gamma_2$ and have  intervals $[1,4)$ and $[2, 3)$, respectively. The single image-persistent cycle of $\PH_1(g)$ is generated by $\delta$ and  its interval is $[0, 3)$. Hence, by Definition \ref{defn:match_pd} we have $\gamma_2\overset{\mathcal{F}_{Z,\Int}}{\sim}\delta$.

\begin{remark}
The matching of intervals 
in Definition \ref{defn:match_pd}, is  similar to the barcode matchings scheme presented in \cite{bauer_ind_match_2013}. There, the authors define an interval matching map between barcodes of two persistence modules $M$ and $N$  connected by a morphism $f:M\rightarrow N$. They state that the surjective map $M\rightarrow\Imm f$ induces a surjective map between the barcodes $q:\mathcal{B}(M)\rightarrow\mathcal{B}(\Imm f)$ on birth times, i.e.~an interval $[b,d)\in\mathcal{B}(M)$ is mapped to an interval $[b,d')\in\mathcal{B}(\Imm f)$ with the same birth time. Similarly, the injective map $\Imm f\rightarrow N$ induces an injective map $j:\mathcal{B}(\Imm f)\hookrightarrow \mathcal{B}(N)$ on death times. Then, the composition $h=j\circ q$ is a matching between the intervals of $\mathcal{B}(M)$ and $\mathcal{B}(N)$.
Going back to our setting, taking the morphisms $f_*:\PH_k(X)\rightarrow\PH_k(Z)$ and $g_*:\PH_k(Y)\rightarrow\PH_k(Z)$, 
Definition \ref{defn:match_pd} can be restated in the following way.
The intervals $\gamma\in\PH_k(X)$ and $\delta\in\PH_k(Y)$ are matching intervals if $h_f(\gamma)=h_g(\delta)$, where $h_f$ and $h_g$ are the corresponding induced matchings of $f_*$ and $g_*$ between barcodes, as defined by \cite{bauer_ind_match_2013} (where, abusing notation, we use $\gamma,\delta$ to denote both the persistent cycles and their corresponding barcode intervals). 
\end{remark}

\section{Cycle Registration for Simplicial Complexes}\label{sec:alg}
The definitions in Section \ref{sec:registration} apply to any topological spaces, with a proper notion of homology. In this section we want to focus on the case where the topological spaces are simplicial complexes, and provide the required algorithms that implement our cycle matching framework.

We start with a brief description of the computation of image-persistent homology presented in \cite{cohen_steiner_pers_image_2009}. 
\subsection{Computing image-persistent homology}
Let $X$ and $Z$ be finite simplicial complexes associated with the filtrations $\mathcal{F}_Z=\{Z_m\}_{m=0}^M$ and $\mathcal{F}_X=\{X_m\}_{m=1}^M$ respectively. 
Let $f_m:X_m\rightarrow Z_m$ be simplicial maps such that  $f_j|_{X_i}=f_i$ for $0\leq i\leq j\leq M$, implying that $\Imm f_i\subset\Imm f_j$. 
Thus, the sequence $\{\Imm f_m\}_{m=1}^M$ is a filtration of simplicial complexes. 
Note that the discrete nature of simplicial complexes allows us to assume, without loss of generality, that the filtration index is discrete as well. 
Throughout this section, we will also assume that the maps $f_m$ are
injective, since otherwise chains in $C_k(X)$ that are not necessarily $k$-cycles can be mapped to $k$-cycles in $Z$. 

Computing the image-persistent homology $\PH_k(f)$ 
follows the same steps as computing the standard persistent homology of $\mathcal{F}_Z$ but with a little tweak. 
First, the boundary matrix for $\mathcal{F}_Z$ is generated. 
Then, the rows of the matrix are reordered in the following way. Rows that refer to simplexes in $\Imm f_M$, are sorted according to the order of their pre-image in $\mathcal{F}_X$. Then,  all other rows that refer to $k$-simplexes in $Z\setminus\Imm f$, are pushed to the end of the matrix, while preserving their inner order, after the rows that correspond to the $k$-simplexes of $\Imm f$.
Next, the standard reduction procedure is performed, while the $k$-cycles that are generated by simplexes in $Z\setminus\Imm f$ are discarded. For more details and a proof for the correctness of this algorithm see \cite{cohen_steiner_pers_image_2009}.

\begin{figure}
    \centering
    \includegraphics[width=0.75\textwidth]{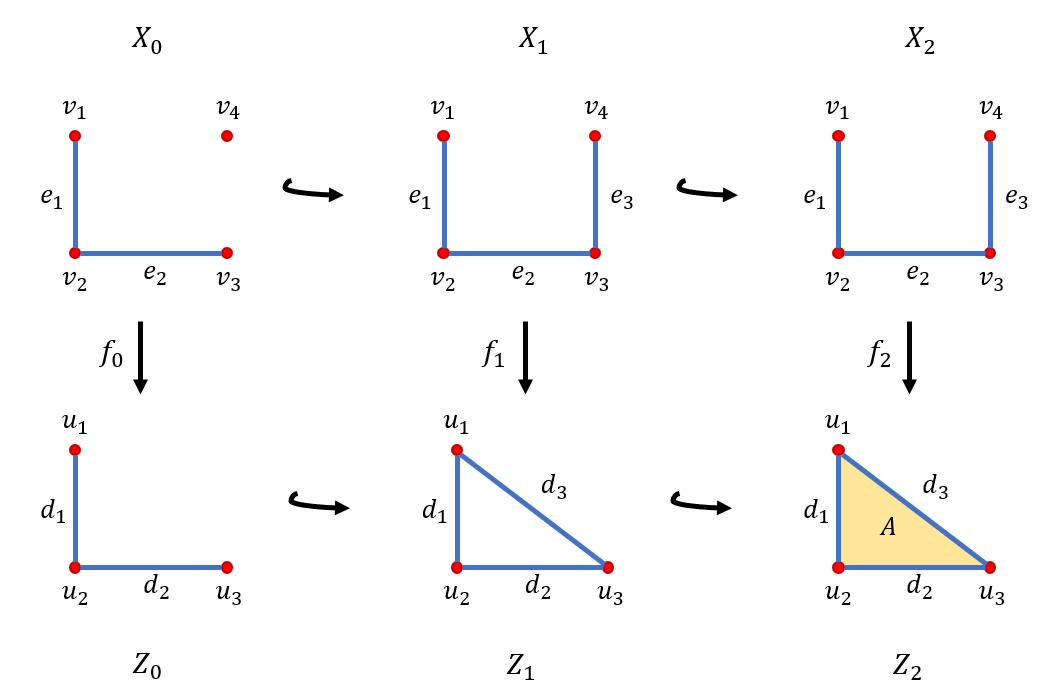}
    \caption{Image-persistence algorithm fail. The filtration $X$ in the top row is mapped to the filtration $Z$ in the bottom row by the simplicial maps $f_0,f_1$ and $f_2$, induced by mapping $v_i\rightarrow u_i$ for $i=1,2,3$ and $v_4\rightarrow u_1$. There is no persistent cycle in $X$, hence, there is no image-persistent cycle. Since the maps $f_{j}$ are surjective, applying the image-persistence algorithm presented in \cite{cohen_steiner_pers_image_2009} will output a false image-persistent cycle that is born at stage $1$ and dies at stage $2$. This happens since the map is not injective.    }
    \label{fig:alg_fail}
\end{figure}

\begin{remark}
Computing the image-persistent homology when $f$ is not injective, cannot be done using the algorithm presented above as can be seen in the example given in Figure \ref{fig:alg_fail}. An algorithm for the general case will remain as  future work. 

\end{remark}

\subsection{Persistence interval matching}
In this section we explain how to use the image-persistent homology in order to find matching intervals between two filtrations. 
Recall that the setting is that $X$, $Y$ and $Z$ are topological spaces equipped with Morse filtrations $\{X_t\}_{t\in\mathbb{R}},\{Y_t\}_{t\in\mathbb{R}}$ and $\{Z_t\}_{t\in\mathbb{R}}$ such that for all $t$ we have 
$X_t\xrightarrow{f_t} Z_t\xleftarrow{g_t} Y_t$.
In order to find a match between generators of intervals in $X$ and $Y$, we compute image-persistent homology twice for both $f$ and $g$. Then, we find a match between the generators of the image-persistent homologies by comparing their death times. If a cycle in $\PH_k(f)$ and a cycle in $\PH_k(g)$ have the same death time, then they refer to the same homological feature in $Z$.
The case where $f_t,g_t$ are injective maps is given below in Algorithm \ref{alg:int_match}.
Note that, the function \textbf{reduceImage} in lines 1,2 of Algorithm \ref{alg:int_match} applies the algorithm presented in \cite{cohen_steiner_pers_image_2009} and returns all finite intervals of the $k$-th image-persistence, in a form of a dictionary that pairs positive simplexes indices with their corresponding negative simplexes indices.

\begin{algorithm}
 	\caption{Function imageCycleMatching}
 	\label{alg:int_match}
	\SetKwInOut{Input}{input}\SetKwInOut{Output}{output}
	\Input{$\partial_{k+1}^Z$ - the $(k+1)$-th boundary matrix of $Z$. \\
	$F_k^X,F_k^Y$ -  filtration values of the $k$-simplexes of $X$ and $Y$.\\
    $F_{k+1}^Z$ - filtration values of the $(k+1)$-simplexes of $Z$.\\ 
	$D_k^X,D_k^Y$ - dictionaries that map simplexes in $X,Y$ to simplexes in $Z$. \\
	} 
	\Output{$CC$ - the pairs of matching intervals between $\PH_k(X)$ and $\PH_k(Y)$.}	
\BlankLine
	$L_{XZ}\leftarrow$ \textbf{reduceImage}($\partial_{k+1}^Z,F^X_{k},F^Z_{k+1},D_k^X$)\;
	$L_{YZ}\leftarrow$ \textbf{reduceImage}($\partial_{k+1}^Z,F^Y_{k},F^Z_{k+1},D_k^Y$)\;
	$CC\leftarrow\emptyset$\;
	\For{every $i$ in $L_{XZ}$}{
		\For{every $j$ in $L_{YZ}$}{
			\If{$L_{XZ}[i]==L_{YZ}[j]$}{
				$CC\leftarrow CC\cup (i,j)$\;
			}
		}
	}
	\textbf{return} $CC$\;
	
\end{algorithm}

\paragraph{Infinite intervals.}
The persistence-interval matching algorithm is based on matching image-persistent cycles that share the same negative simplex (see Definition \ref{defn:match_pd}). In other words, the matching process involves only finite intervals, since infinite intervals do not have negative simplexes associated with them. This poses a problem when the spaces under consideration have infinite intervals that we might want to match.

In order for Algorithm \ref{alg:int_match} to handle infinite persistence intervals as well, we make all the infinite intervals ``pseudo-finite'', by filling each infinite cycle as follows. Let $X,Y$ and $Z$ as before. We take one of the spaces, without loss of generality $X$, and list all cycles that generate infinite persistence intervals, $\gamma_1,\ldots,\gamma_n$. For each $\gamma_i$ we compute its image $f(\gamma_i)$ and assign a pseudo eliminating chain $c_i$, such that $\partial c_i=f(\gamma_i)$.  
Then, we concatenate $c_1,\ldots,c_n$ corresponding columns to the end of the matrix $\partial_{k+1}^Z$ 
after all its original $(k+1)$-simplexes. In order to assure that the infinite intervals are considered as legal intervals (i.e.~the filtration value of the negative simplex is higher than the value of the positive simplex) we assign these columns with infinite filtration values.
We can then apply Algorithm \ref{alg:int_match} to the extended matrix. We say that two infinite intervals match if they are eliminated by the same pseudo chain.

Note that in general the problem of cycle registration in the fixed homology setting (see Definition \ref{defn:match_fixed}), can be interpreted as matching infinite intervals in the following sense. The fixed homology can be thought of as the persistent homology of a filtration of length one. This way, fixed homology cycles identify with the infinite persistence cycles (which are the only persistent cycles in that case). Hence, the algorithms described above work in the fixed homology setting as well.

\section{Applications}\label{sec:app}
\subsection{Homology Inference}\label{sec:hom_inf}

\begin{figure}
    \centering
    \includegraphics[width=0.5\textwidth]{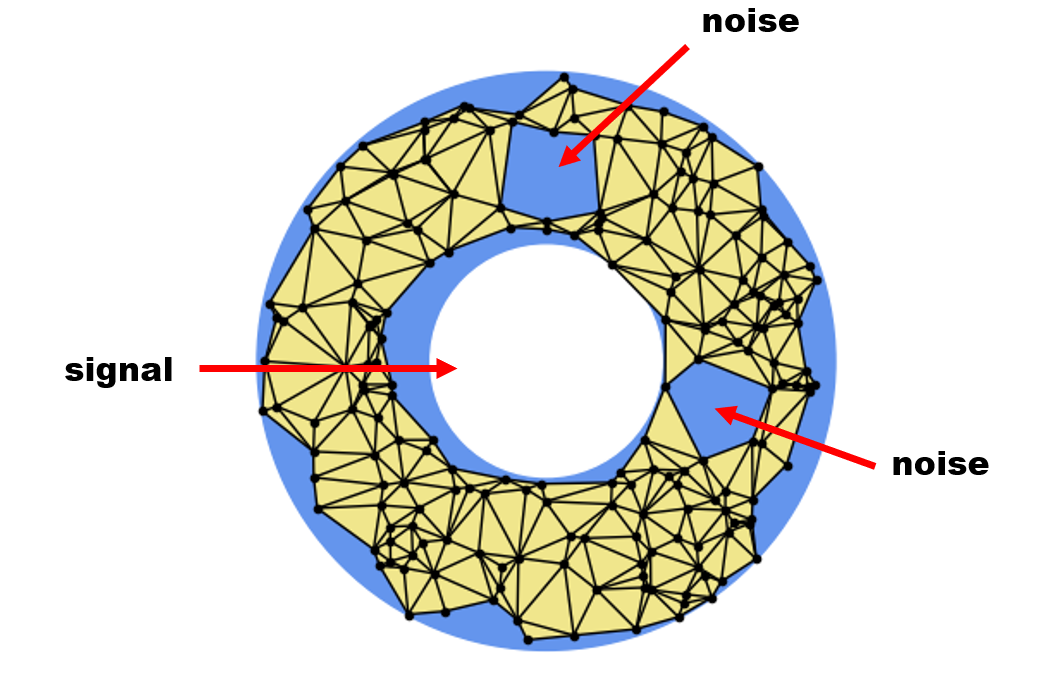}
    \caption{Differentiating signal from noise. A simplicial complex over a dataset sampled uniformly from an annulus (blue). The central hole in the complex corresponds to the hole of the annulus (signal), while the other two holes are artifacts (noise).}
    \label{fig:signal_noise}
\end{figure}

Recall the following motivational question presented in the introduction. In the course of uncovering the underlying homology of a given dataset,
how can we differentiate between \emph{signal} (i.e.~true features of the data) and \emph{noise}? See example in Figure \ref{fig:signal_noise}.
In this section we address this question by suggesting an approach that utilizes the cycle registration method presented in the previous chapters.

\paragraph{Geometric complexes.}
Let $\cX,\cY\subset \R^d$ be finite sets, and let $X = \Cech_r(\cX), Y=\Cech_r(\cY)$ be the corresponding \cech complexes (see Definition \ref{def:cech}).
The cycle registration scheme described in Sections \ref{sec:cyc_match} and \ref{sec:ph_match} can be applied naturally for this choice of $X$ and $Y$, by taking $Z = \Cech_r(\cX\cup \cY)$ with $f,g$ taken to be the inclusion maps, since  
$$\Cech_r(\cX)\subset\Cech_r(\cX\cup\cY)\supset \Cech_r(\cY).$$
The same applies to the Rips complex (see Definition \ref{def:rips}).

\begin{remark}
A significant drawback of using either the \cech or the Rips in applications is the size of the complexes generated. This problem is exacerbated when applying the cycle registration scheme presented in this paper.
In \cite{coupled_alpha} we introduce a new geometric complex we named `the coupled alpha complex' that is based on the well-known alpha complex \cite{edelsbrunner_alpha_shapes}, and allows us to preform cycle registration for the \cech complex in a significantly more efficient way. In fact, we use this complex in the simulations presented below.
\end{remark}

\subsubsection{Statistical bootstrap}
In real data analysis applications, the underlying distribution of a given sample is usually unknown. Hence, explicitly computing the properties of an estimator (e.g.~variance) becomes impossible.
Roughly speaking, the term \emph{statistical bootstrap} \cite{efron_introduction_1994} refers to methods that approximate statistical properties of estimators using the same data from which the estimator is calculated. 
The idea of bootstrap is to study the properties of an estimator using an approximated distribution (e.g.~the empirical distribution of the sample). 
The power of bootstrap comes from the fact that it does not require additional information or data, as well as from its simplicity - approximating the properties of an estimator is done in a straightforward way even for complex estimators. We demonstrate this in the following example.

\paragraph{Example.} Let $\cX=\{X_1,\ldots,X_n\}\subset\mathbb{R}$ be a set of  i.i.d.~random variables sampled from an unknown distribution with mean value $\mu$.
The standard unbiased estimator for $\mu$ is 
$$\hat{X}=\dfrac{1}{n}\sum_{i=1}^n X_i.$$
In order to assess the accuracy of $\hat{X}$, one might want to know its variance. The bootstrap framework allow us to
approximate the variance of $\hat{X}$ in the following way. First, generate additional samples $\cX^{(1)},\ldots,\cX^{(B)}$, by repeatedly sampling $n$ points from $\cX$ with replacement.
Next, compute the empirical mean for each of the resamples,
$$\hat{X}^{(j)}=\dfrac{1}{n}\sum_{i=1}^n X^{(j)}_i,\quad\forall j\in\{1,\ldots,B\}.$$
The approximated variance is then given by
$$\hat{\sigma^2}=\dfrac{1}{n}\sum_{j=1}^n \|\hat{X}^{(j)}-\hat{X}\|^2.$$

It can be shown that under mild assumptions the bootstrap approximation converges to the real variance 
as the number of points increase \cite{efron_introduction_1994}.
In other words, we use the empirical distribution as a proxy for the real distribution, in order to approximate the properties of the given estimator. 

\subsubsection{Cycle prevalence}
As we described in the introduction, the main motivation for this work is to develop a bootstrap-like framework that will enable us to quantify the likelihood of an observed topological feature to be representing information about the underlying distribution (the signal). 

\begin{figure}
    \centering
    \begin{subfigure}{0.32\textwidth}
        \includegraphics[width=\textwidth]{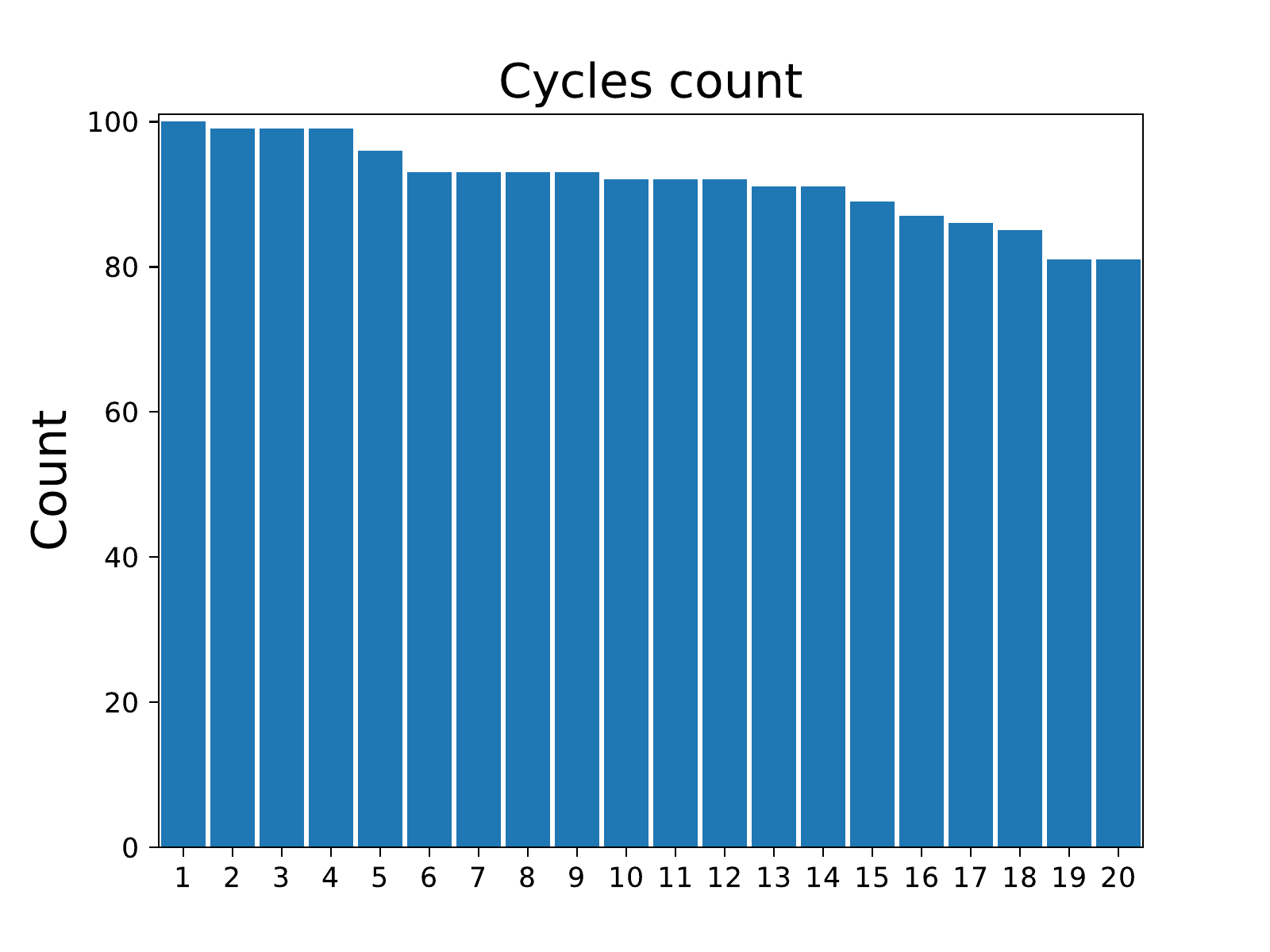}
    \end{subfigure}
    \begin{subfigure}{0.32\textwidth}
        \includegraphics[width=\textwidth]{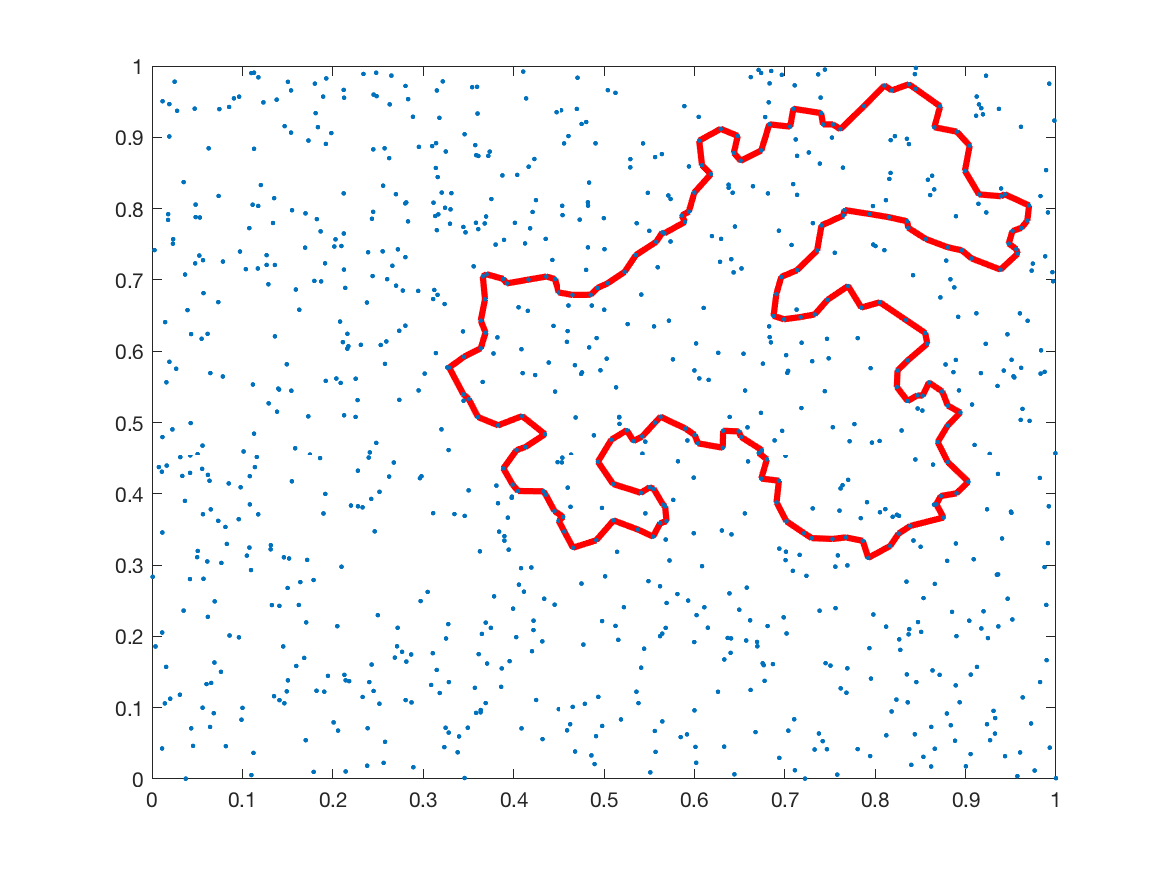}
    \end{subfigure}
    \begin{subfigure}{0.32\textwidth}
        \includegraphics[width=\textwidth]{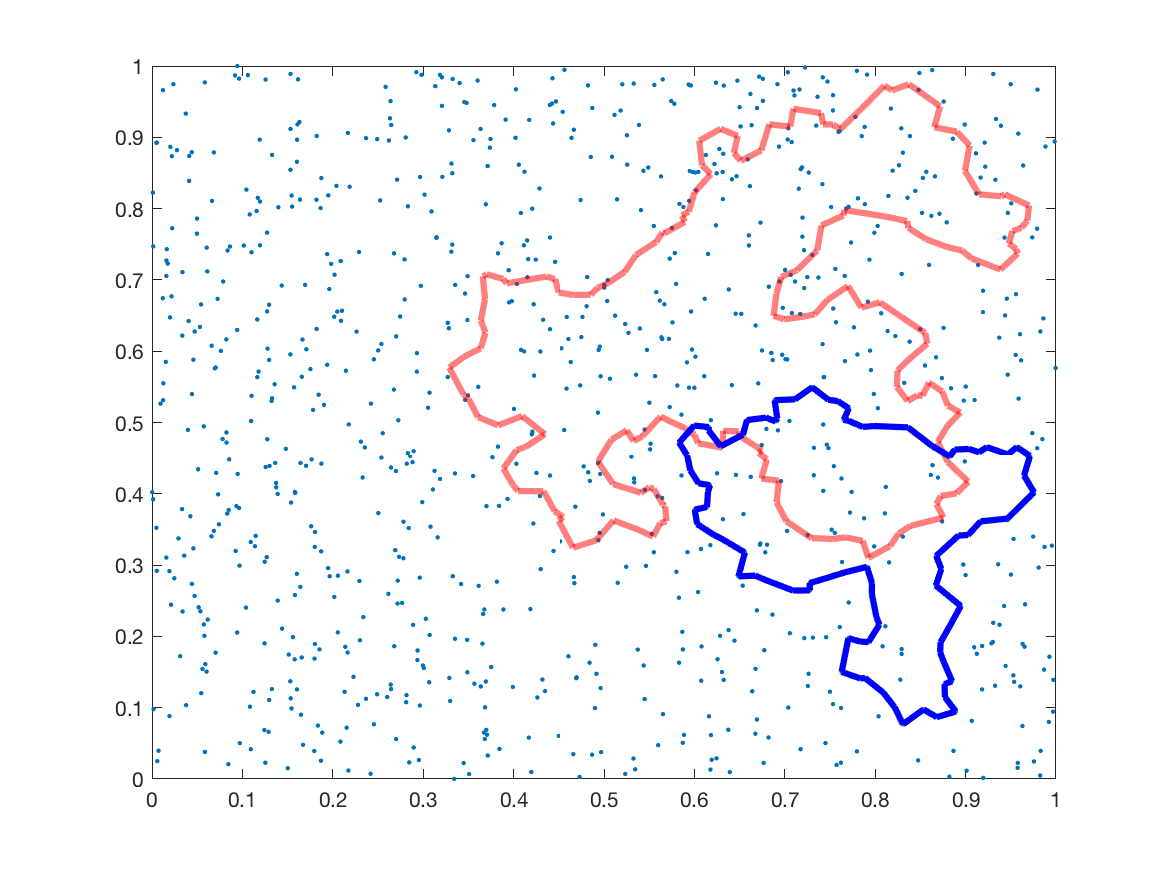}
    \end{subfigure}
    \begin{subfigure}{0.32\textwidth}
        \includegraphics[width=\textwidth]{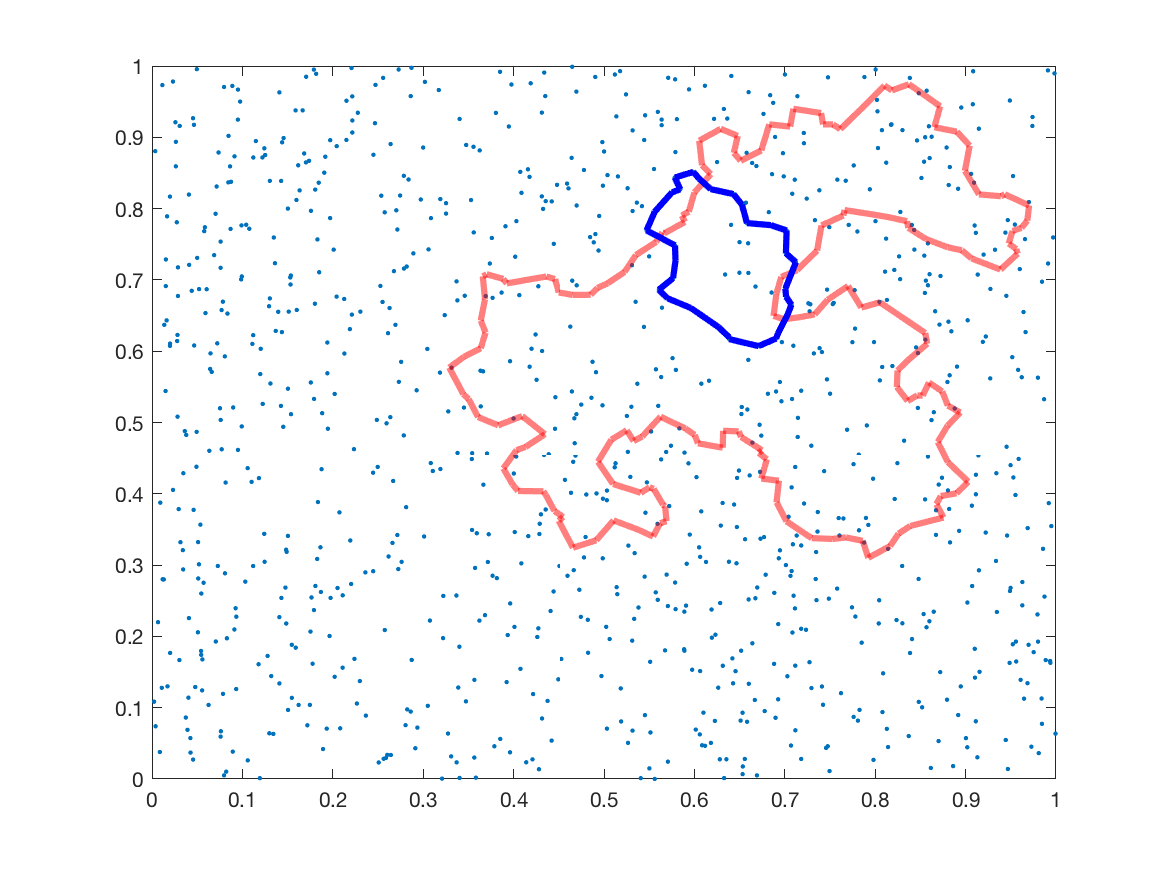}
    \end{subfigure}
    \begin{subfigure}{0.32\textwidth}
        \includegraphics[width=\textwidth]{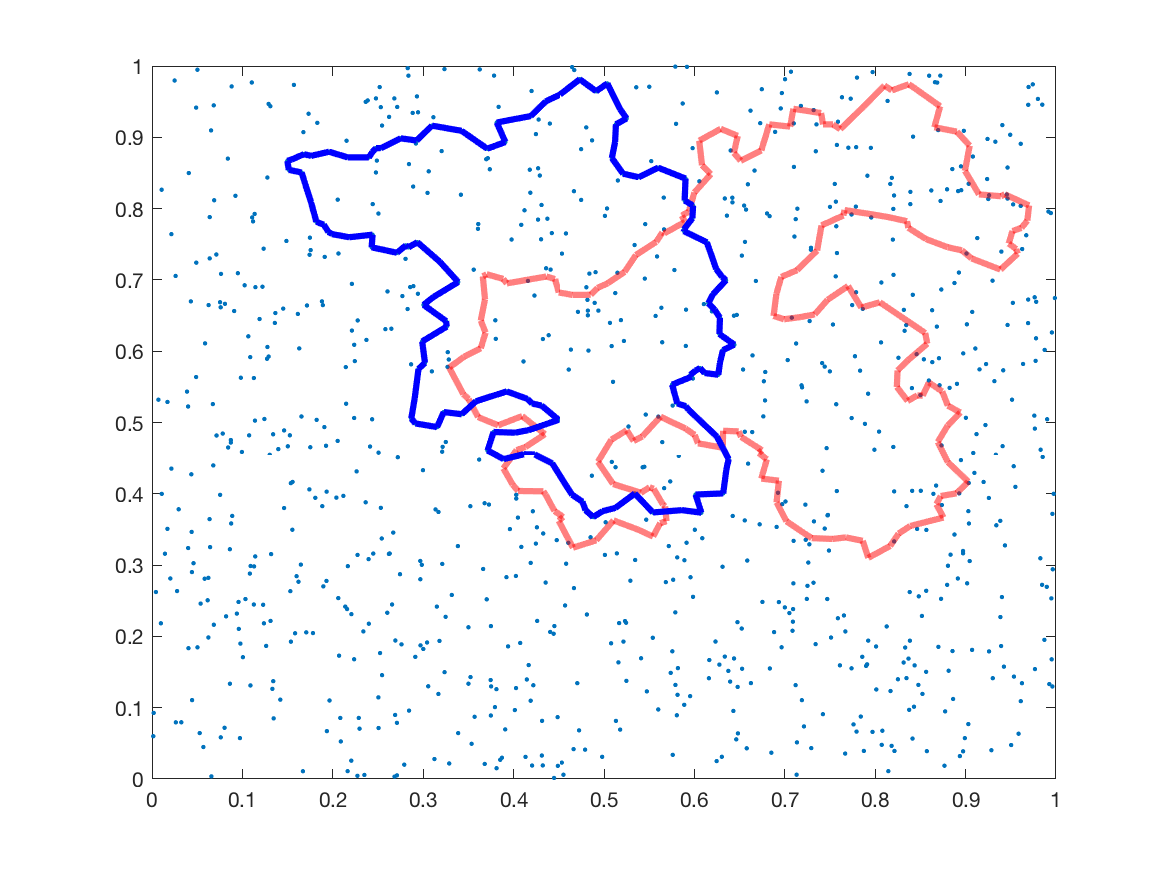}
    \end{subfigure}
    \begin{subfigure}{0.32\textwidth}
        \includegraphics[width=\textwidth]{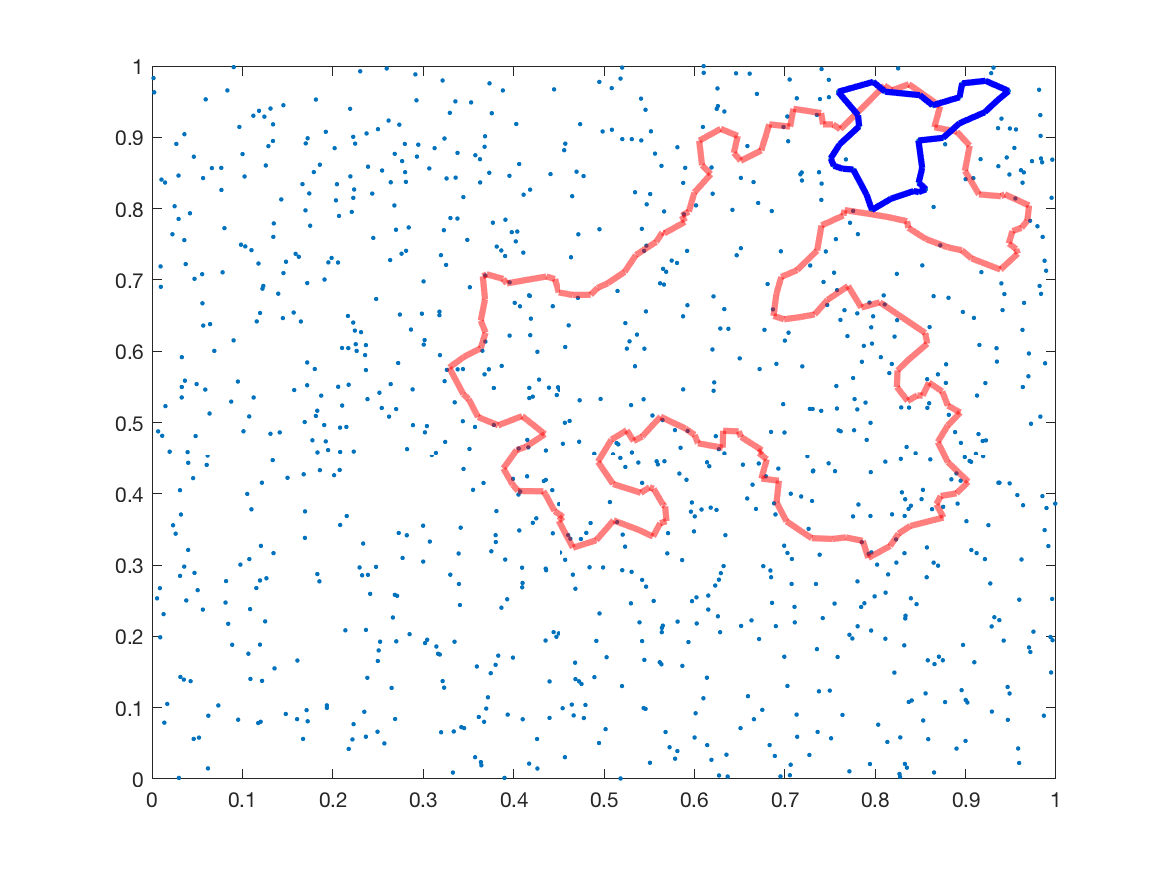}
    \end{subfigure}
    \caption{Noisy cycle generated by sampling $1000$ points from the uniform distribution on the unit square and its matches in additional samples. Top left: The cycles count for the twenty cycles in the original sample with the highest number of matches in $100$ additional samples. Red: the cycle with the highest percentage of matches. Blue: matching cycles from different additional samples. The blue cycles have inconsistent characteristics. }
    \label{fig:noise_problem}
\end{figure}

Our initial proposal was to represent this likelihood using the probability that an observed persistent cycle will reappear in a new sample.
However, while experimenting with data, we noticed that as the sample size grows noisy generators tend to reappear frequently  (see Figure \ref{fig:noise_problem}), making it impossible to differentiate signal from noise. 
This phenomenon occurs due to the fact that large noisy cycles, i.e. cycles whose generators composed of a large number of simplexes, are more likely to appear in large samples.
As can be seen in Figure \ref{fig:noise_problem}, each of these cycles encloses a large amount of `volume'. Hence, they are more likely to have overlapping cycles in an additional sample, which in turn results in a greater chance for a match. 

Since frequency alone is not enough, we decided to estimate a finer measure that in addition takes into account the lifetimes of both persistent cycles and image-persistent cycles as follows.

\begin{definition}
Let $X,Y$ and $Z$ be topological spaces equipped with Morse filtrations $\{{X}_t\}_{t\in\mathbb{R}}$, $\{{Y}_t\}_{t\in\mathbb{R}}$, and $\{{Z}_t\}_{t\in\mathbb{R}}$ respectively,  such that for all $t$ we have
${X}_t\xrightarrow{f_t} {Z}_t\xleftarrow{g_t} {Y}_t$. \\
Let $\gamma\in\PH_k(X)$ and $\delta\in\PH_k(Y)$ be such that $\gamma\overset{\mathcal{F}_{Z,\Int}}{\sim}\delta$, and let ${\tilde\gamma}\in\PH_k(f)$ and ${\tilde\delta}\in\PH_k(g)$ be their corresponding image-persistent cycles. We define the following ratios
$$a_\gamma = \dfrac{|\Int(\gamma)\cap\Int({\tilde\gamma})|}{|\Int(\gamma)\cup\Int(\tilde\gamma)|},\quad a_\delta = \dfrac{|\Int(\delta)\cap\Int({\tilde\delta})|}{|\Int(\delta)\cup\Int({\tilde\delta})|},\quad\text{and}$$
$$c_{\gamma,\delta} = \dfrac{|\Int(\gamma)\cap\Int(\delta)|}{|\Int(\gamma)\cup\Int(\delta)|},$$
where $\Int(\cdot)=[\bth(\cdot),\dth(\cdot))$ denotes the lifetime interval, and $|\cdot|$ denotes interval  length.
\end{definition} 
Note that all these ratios yield numbers between $0$ and $1$.
The values of $a_\gamma, a_\delta$ measure the similarity between a cycle and its corresponding image-persistent cycle in terms of lifetime lengths.
The value of $c_{\gamma,\delta}$ measures the amount of overlap between the matching cycles lifetimes.
Next, we define the following \emph{cycles affinity} measure, 
$$\rho_{\gamma,\delta}:=a_\gamma a_\delta c_{\gamma,\delta}.$$
This measure quantifies the strength of the matching between cycles. In other words, a match is considered strong if the cycles have overlapping lifetimes with each other and with their corresponding image-persistent cycles.

\subsubsection{Simulation results}
\paragraph{Inference from additional samples.}
Let $\cX=\{X_1,\ldots,X_n\}\subset\mathbb{R}^d$ be a set of i.i.d.~random variables sampled from a distribution with density function $f:\mathbb{R}^d\rightarrow\mathbb{R}$.  
Let $\mathcal{F}_\cX$ be its \cech filtration and  $\PH_k(\cX)$ the corresponding $k$-th persistent homology.
We generate additional independent samples of size $n$ from $f$, denoted by $\cX_1,\ldots,\cX_B$,
and define $\cZ_i=\cX\cup\cX_i$. 
Take any $\gamma\in\PH_k(\cX)$, and denote by $\delta_i\in\PH_k(\cX_i)$ its matching cycle via $\mathcal{F}_{\cZ_i}$. If such cycle $\delta_i$ does not exist, we set $\rho_{\gamma,\delta_i}=0$. Next, we assign $\gamma$ the following \emph{cycle prevalence} score,
\begin{equation}
\label{eq:cycle_prev}
\hat{p}_\gamma=\dfrac{1}{B}\sum_{i=1}^B \rho_{\gamma,\delta_i}.
\end{equation}
The value of $\hat{p}_\gamma$ takes into account two factors -- the frequency of the cycle and the affinity between $\gamma$ and its matches. 

In reality, the distribution $f$ is usually unknown. Therefore, generating new samples is impossible. In that case we want to bootstrap $p_\gamma$ using resampling, i.e.~generating new samples, denoted $\hat{\cX}^{(1)},\ldots.,\hat{\cX}^{(B)}$, from an empirical distribution. In the examples below we use the following kernel density estimator,
$$\hat{f}_h(x)=\dfrac{1}{N}\sum_{i=1}^N \dfrac{1}{(2\pi)^{d/2}h^{d/2}}e^{-\frac{\|x-X_i\|^2}{2h}}.$$
Sampling from $\hat{f}_h$ is equivalent to sampling from the empirical distribution with the addition of a small Gaussian perturbation, where $h$ controls the variance of perturbation added. 
We can now apply \eqref{eq:cycle_prev} for the new resamples, in order to bootstrap $\hat{p}_\gamma$. 
In the following, we present the results of the suggested scheme for a few test cases.

\paragraph{Torus.} We take the torus as a simple example to demonstrate that our method is consistent with the heuristic approach that points far from the diagonal in the persistence diagram represent real phenomena underlying the data.

In the following examples we sampled $n$ points uniformly on a torus with $r_{\mathrm{center}}=0.3$ and $r_{\mathrm{tube}}=0.2$, where $r_{\mathrm{center}}$ is the radius of the central hole and $r_{\mathrm{tube}}$ is the tube radius. 
Next, we generated additional $100$ samples of size $n$, from the original distribution, and additional $100$ resamples of size $n$, from the kernel density estimator with $h=0.001$. 

\begin{figure}[h]
\begin{multicols}{3}
	\begin{subfigure}{0.3\textwidth}
		\includegraphics[width=\textwidth]{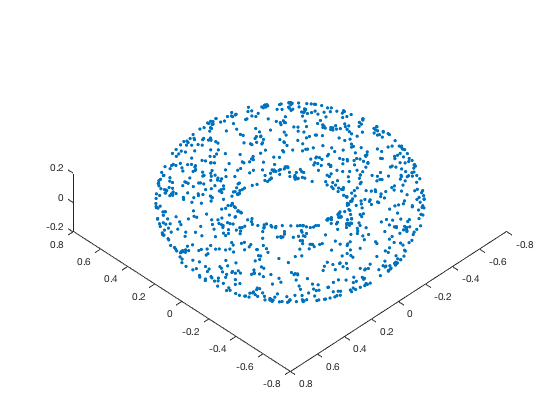}
		\caption{}
	\end{subfigure}
	\begin{subfigure}{0.3\textwidth}
		\includegraphics[width=\textwidth]{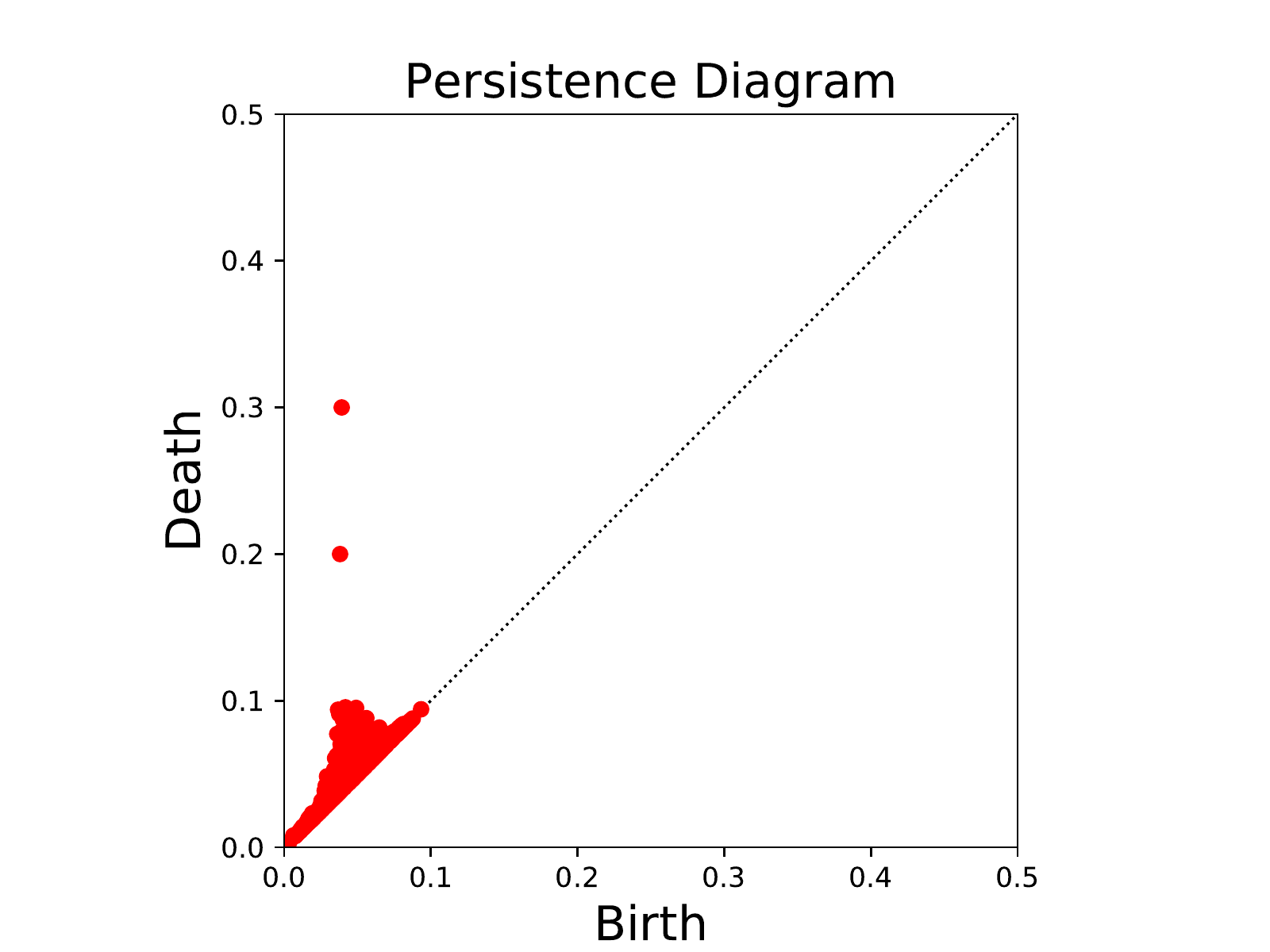}
		\caption{}
	\end{subfigure}
	\begin{subfigure}{0.3\textwidth}
		\includegraphics[width=\textwidth]{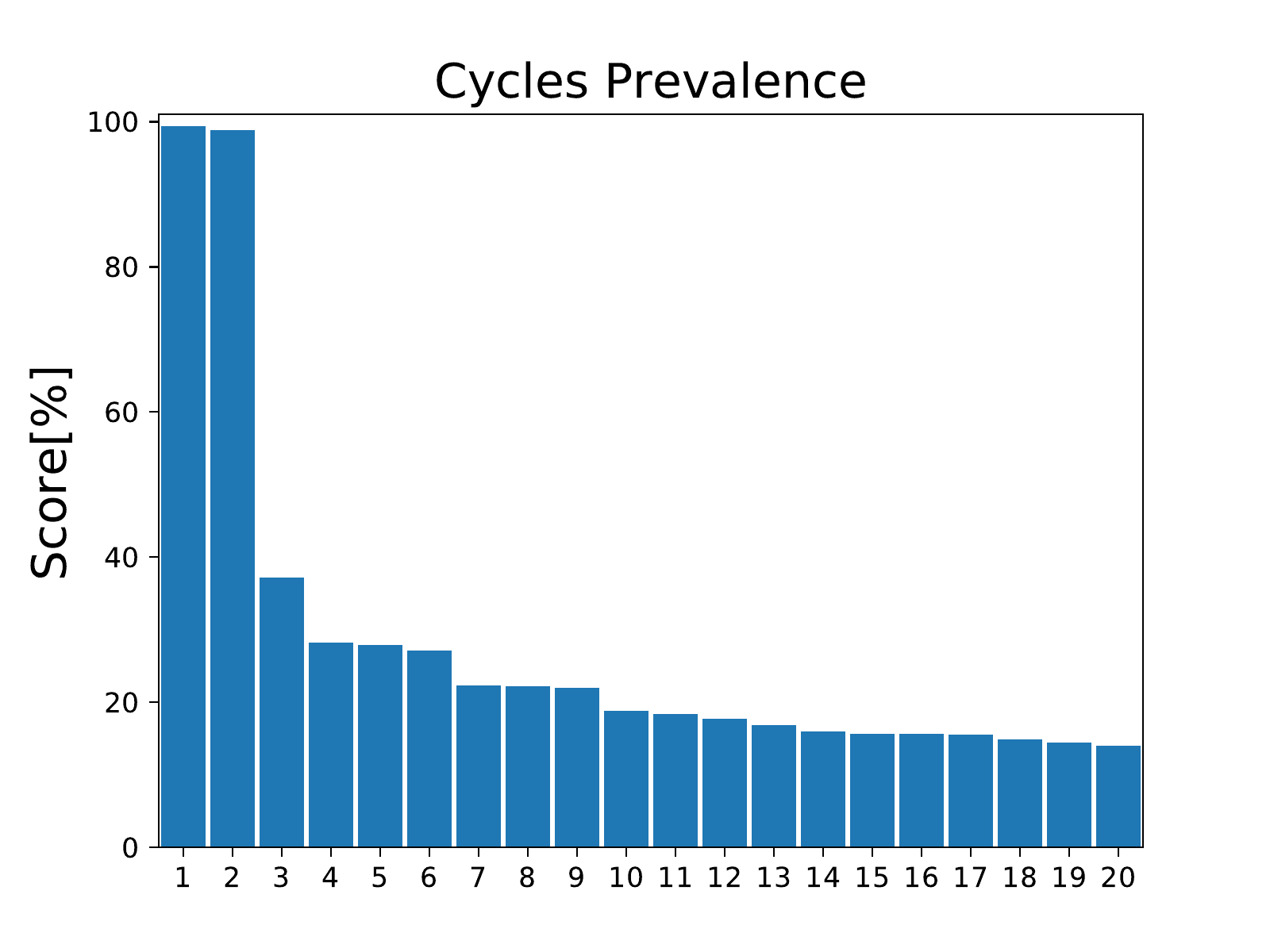}
		\caption{}
	\end{subfigure}	
	\begin{subfigure}{0.3\textwidth}
		\includegraphics[width=\textwidth]{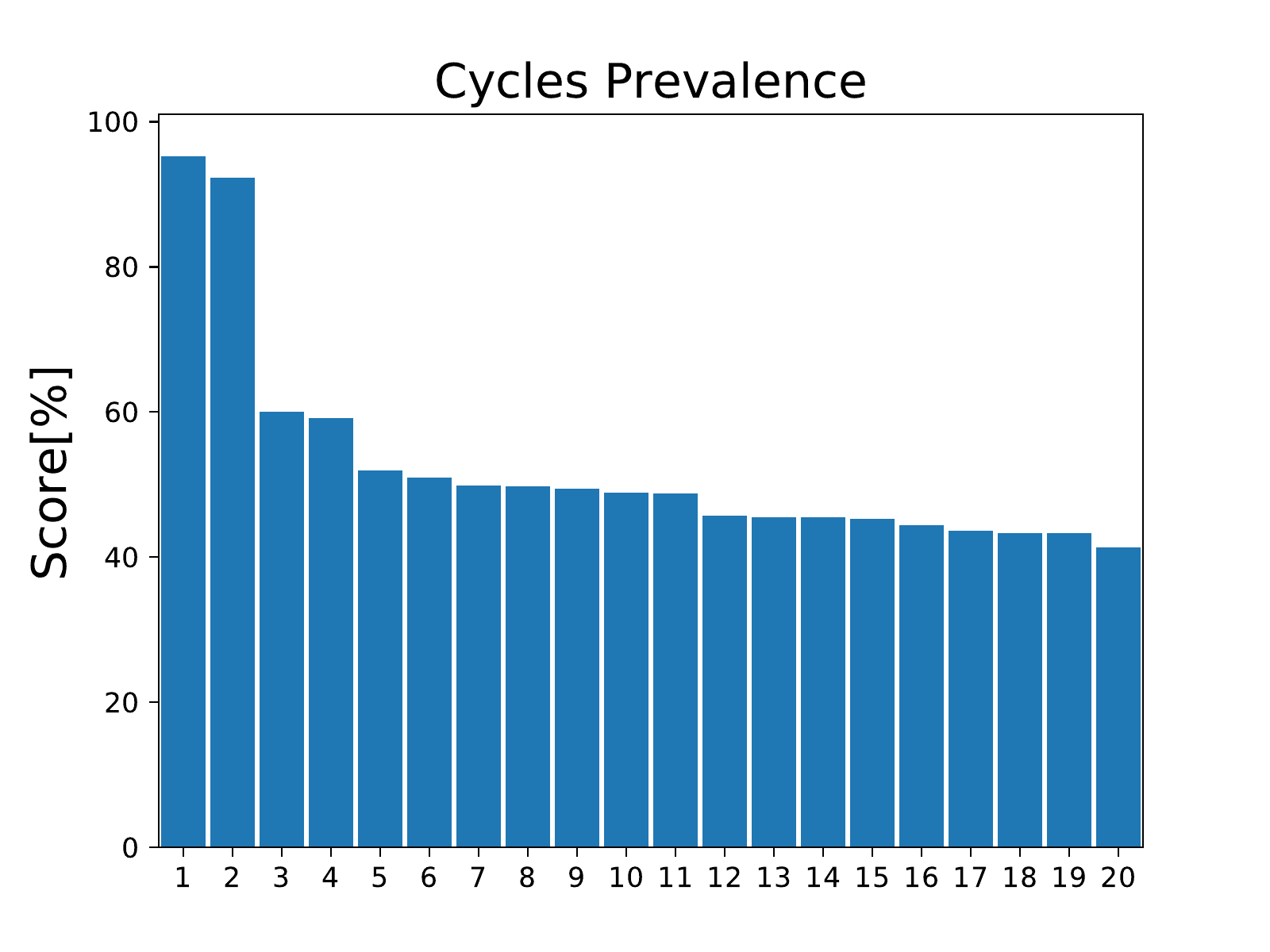}
		\caption{}
	\end{subfigure}
	\begin{subfigure}{0.3\textwidth}
		\includegraphics[width=\textwidth]{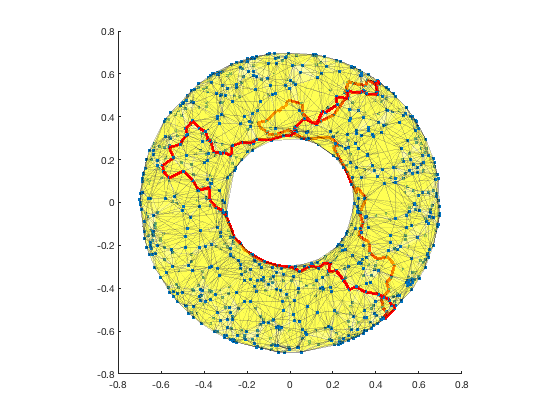}
		\caption{}
	\end{subfigure}	
	\begin{subfigure}{0.3\textwidth}
		\includegraphics[width=\textwidth]{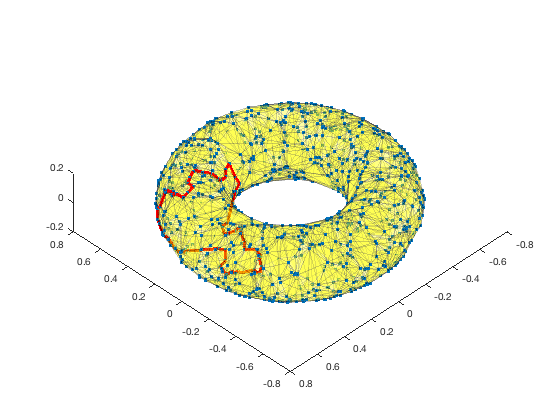}
		\caption{}
	\end{subfigure}
	\end{multicols}
	\caption{Homology inference on the torus. (a) Data -- 1000 points sampled uniformly on a torus, with $r_{\mathrm{center}}=0.3$ and $r_{\mathrm{tube}}=0.2$. (b) The persistence diagram of $\PH_1$. The two points that are furthest from the diagonal correspond to the holes of the torus. (c) The top $20$ cycles in $\PH_1$ with highest prevalence score for new samples. As can be seen there are two prominent cycles. (d)  The top $20$ cycles in $\PH_1$ with highest prevalence score for resamples generated from the kernel density estimator with $h=0.001$. (e) The generator that corresponds to the highest score cycle in both resamples and new samples, viewed from above. (f) The generator that corresponds to the second highest score cycle in both resamples and new samples, viewed from a side angle. It can be seen that it encircles the tube. }
	\label{fig:torus_1000}
\end{figure}

In the first example we took $n=1000$ and computed the cycle prevalence for $\PH_1$ 
in both resamples and in the original distribution samples. The results are presented in Figure \ref{fig:torus_1000}. It can be seen that the cycle prevalence of both resamples and original distribution samples captures the right homology of the underlying torus. 

\begin{figure}[h]
\begin{multicols}{3}
	\begin{subfigure}{0.3\textwidth}
		\includegraphics[width=\textwidth]{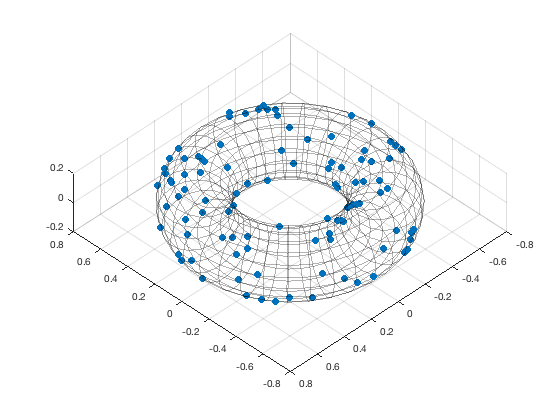}
		\caption{}
	\end{subfigure}
	\begin{subfigure}{0.3\textwidth}
		\includegraphics[width=\textwidth]{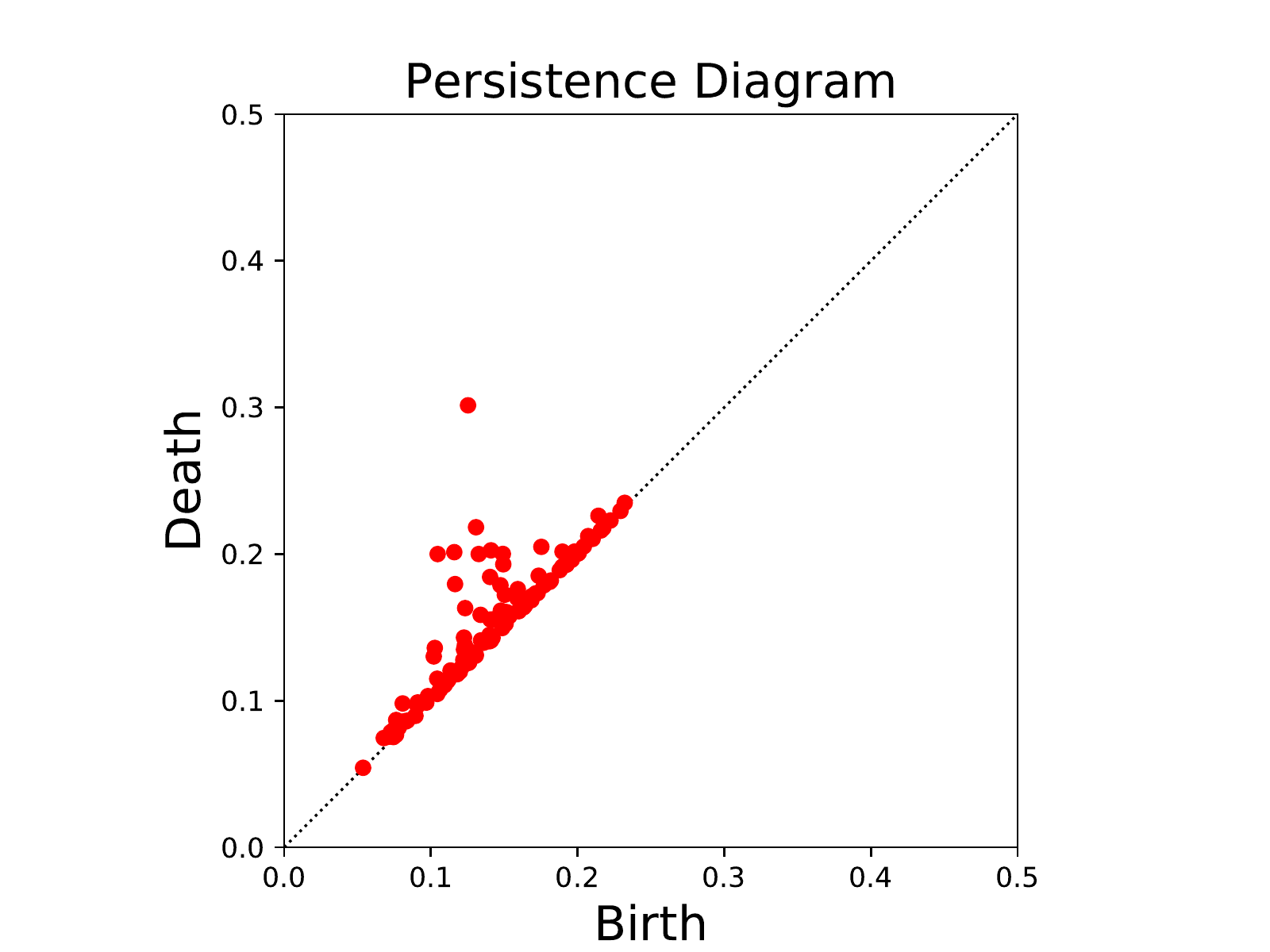}
		\caption{}
	\end{subfigure}
	\begin{subfigure}{0.3\textwidth}
		\includegraphics[width=\textwidth]{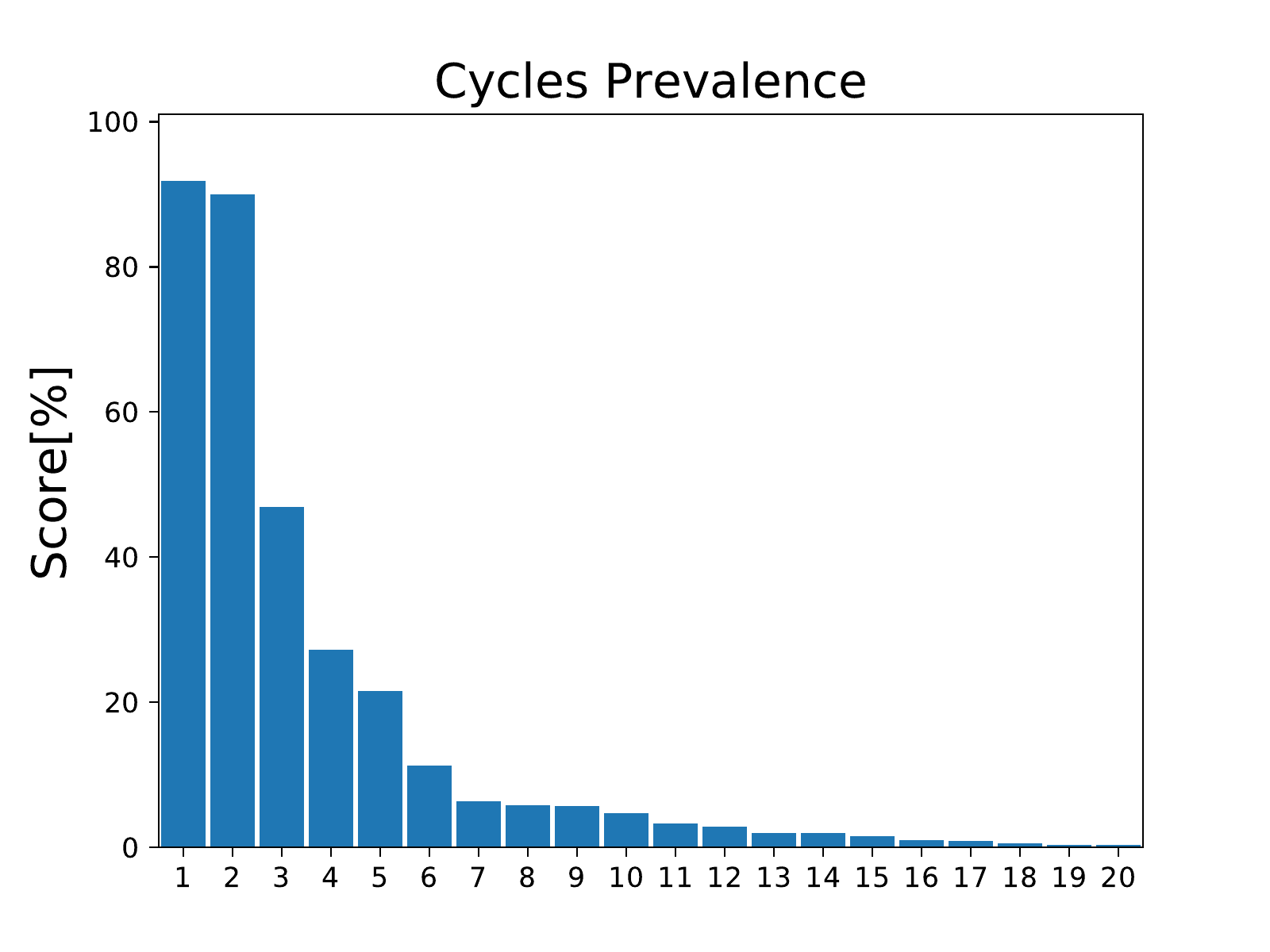}
		\caption{}
	\end{subfigure}
	\begin{subfigure}{0.3\textwidth}
		\includegraphics[width=\textwidth]{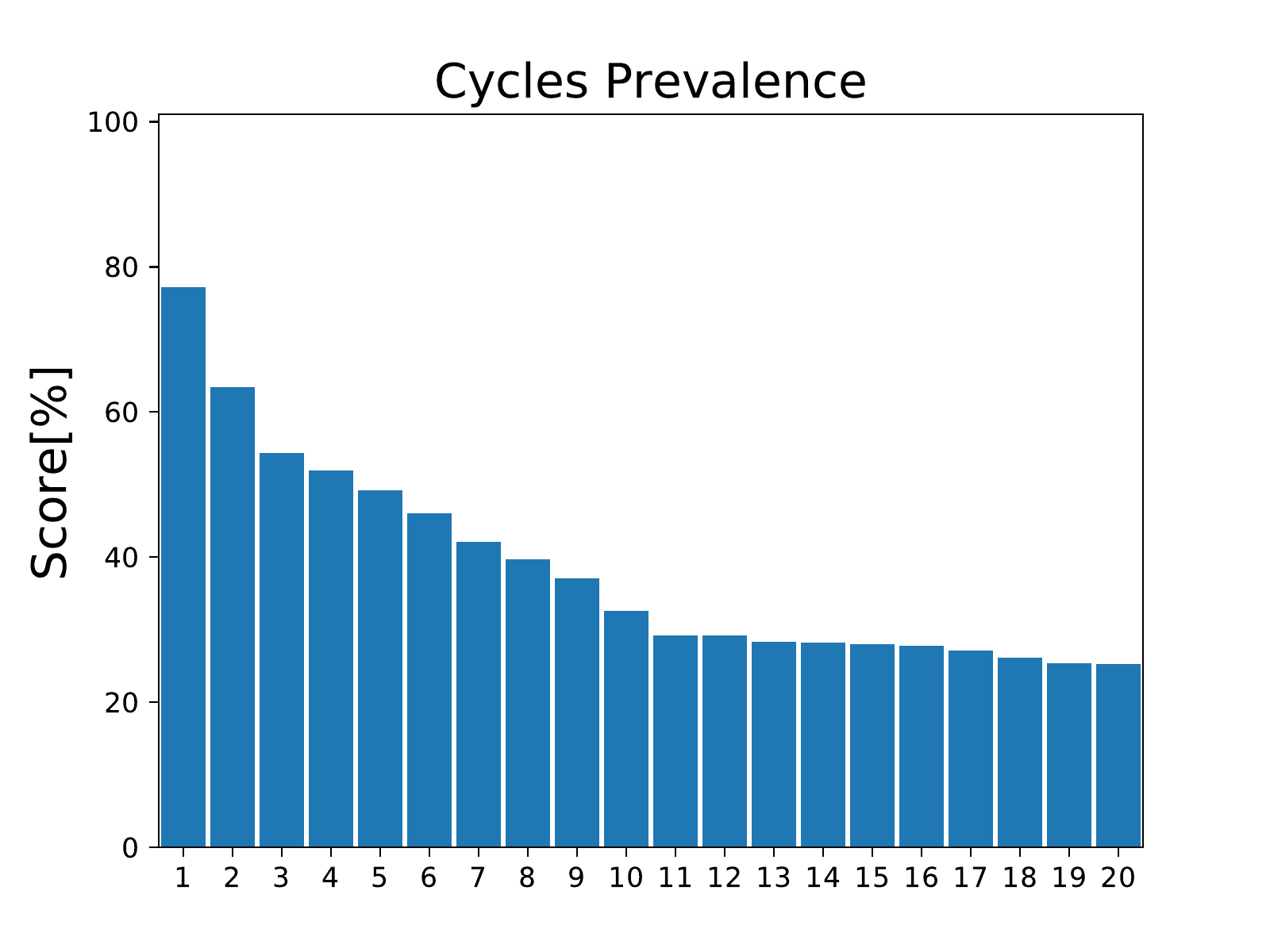}
		\caption{}
	\end{subfigure}
	\begin{subfigure}{0.3\textwidth}
		\includegraphics[width=\textwidth]{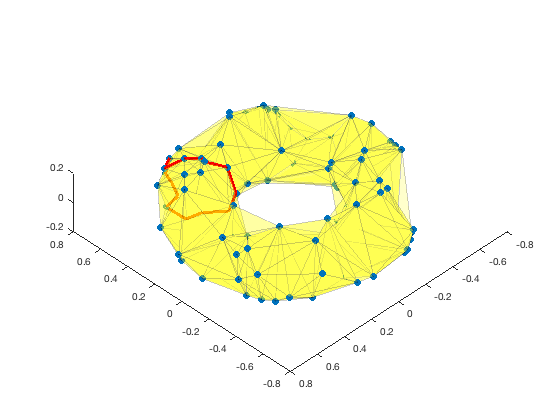}
		\caption{}
	\end{subfigure}
	\begin{subfigure}{0.3\textwidth}
		\includegraphics[width=\textwidth]{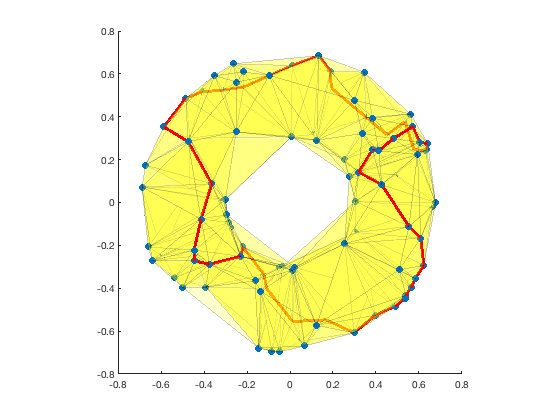}
		\caption{}
	\end{subfigure}
	\end{multicols}
	\caption{Homology inference for the torus. (a) Data sample - 100 points sampled uniformly on a torus, with $r_{\mathrm{center}}=0.3$ and $r_{\mathrm{tube}}=0.2$. (b) The persistence diagram of $\PH_1$. It is hard to identify the point that corresponds to the tube hole. (c) The top $20$ cycles in $\PH_1$ with highest prevalence score for resamples. As can be seen, there are two prominent cycles. (d)  The top $20$ cycles in $\PH_1$ with highest prevalence score for resamples generated from the kernel density estimator with $h=0.001$. (e) The generator that corresponds to the highest score cycle in the new samples, viewed from a side angle. It can be seen that it encircles the tube. (f) The generator that corresponds to the second highest score cycle in the new samples and highest in resamples.}
	\label{fig:torus_100}
\end{figure}

In the second example we took $n=100$ and computed the cycle prevalence for $\PH_1$ in both the resamples and in new samples. The results are presented in Figure \ref{fig:torus_100}.
This time only the cycle that corresponds to the center of the torus is distinguishable and it is very difficult (perhaps impossible) to separate the tube cycle from the noisy cycles in the persistence diagram.
However, in the cycle prevalence figure for new samples the true cycles clearly stand out. Note that the third highest score cycle is also a cycle that encircles the tube. The reason for having more than one cycle of this kind is related to the small sample size. Both cycles are formed due to the tube structure and at least one of them becomes a boundary before their generators become homological. In the resampling setting, the cycle prevalence of the center generator stands out, while the tube hole is less prominent.

\begin{figure}[h]
	\centering
	\begin{subfigure}{0.3\textwidth}
		\includegraphics[width=\textwidth]{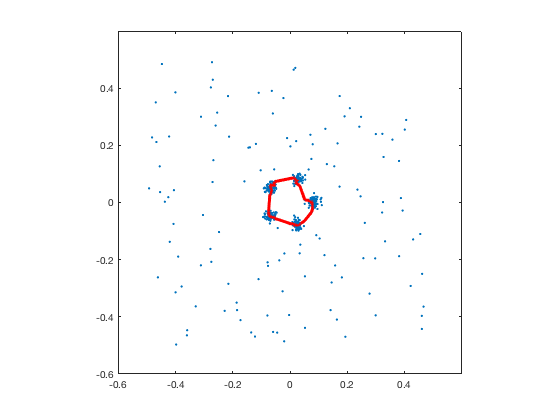}
		\caption{}
	\end{subfigure}
	\begin{subfigure}{0.3\textwidth}
		\includegraphics[width=\textwidth]{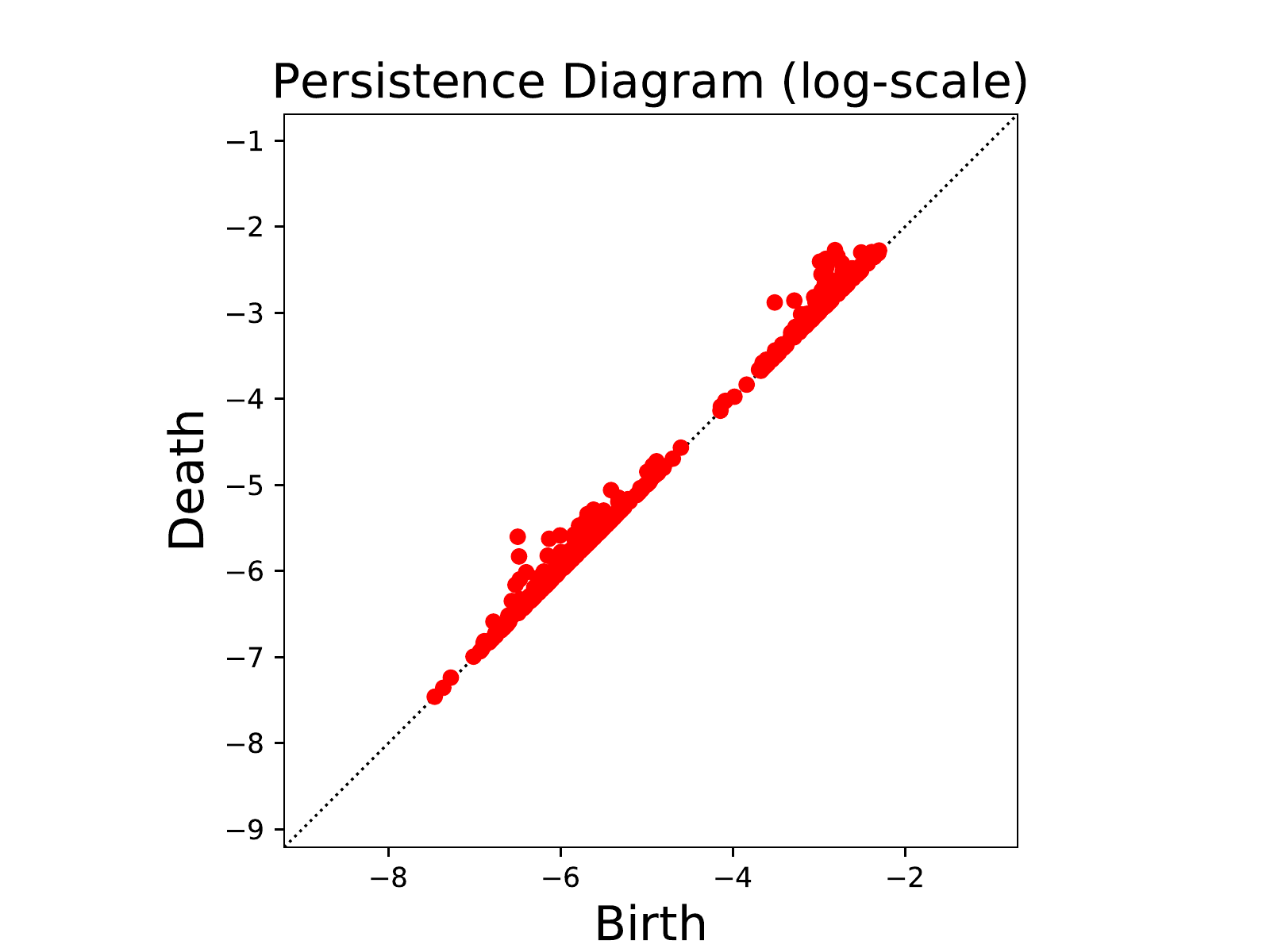}
		\caption{}
	\end{subfigure}
	\begin{subfigure}{0.3\textwidth}
		\includegraphics[width=\textwidth]{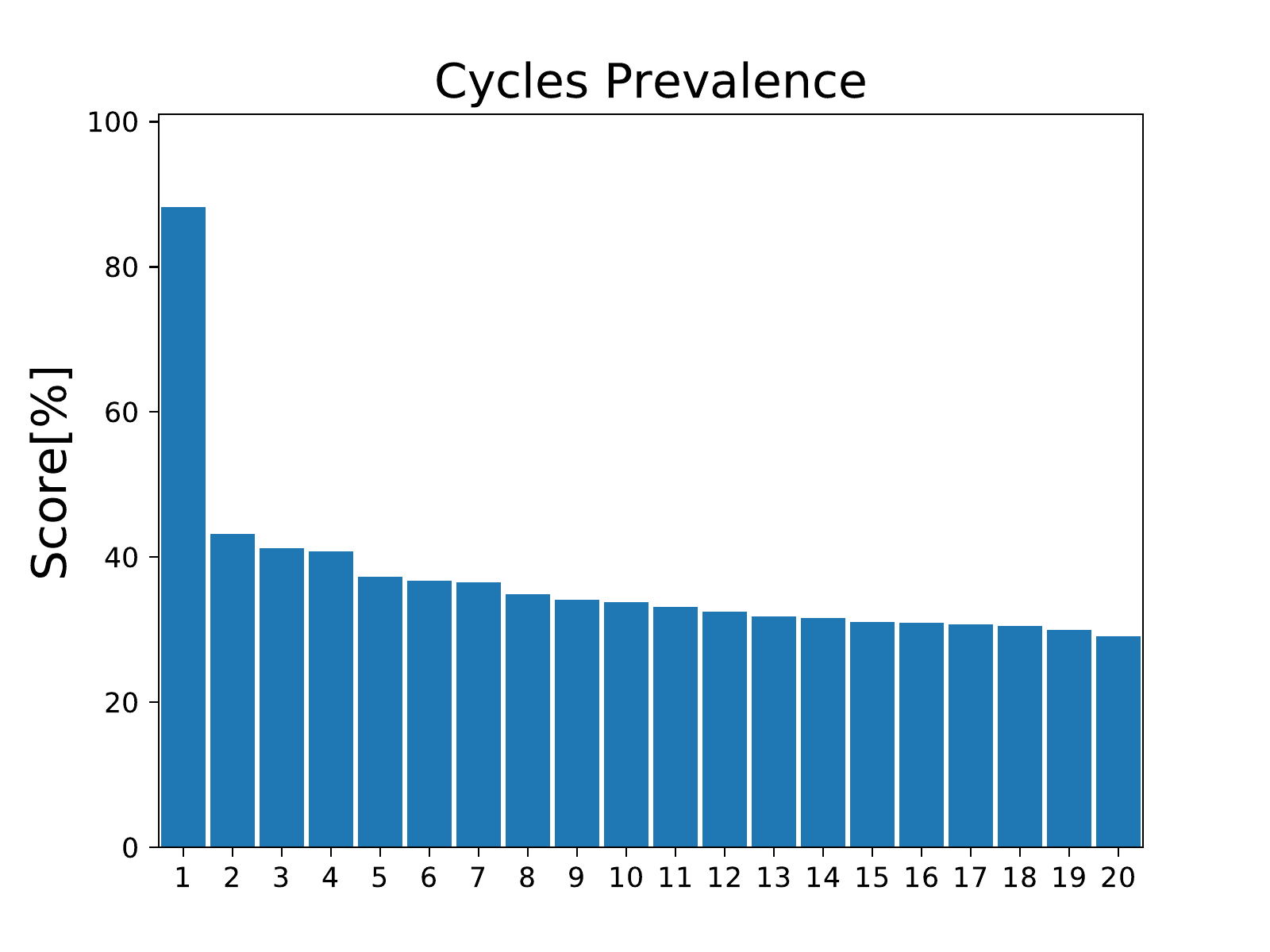}
		\caption{}
	\end{subfigure}
	\begin{subfigure}{0.3\textwidth}
		\includegraphics[width=\textwidth]{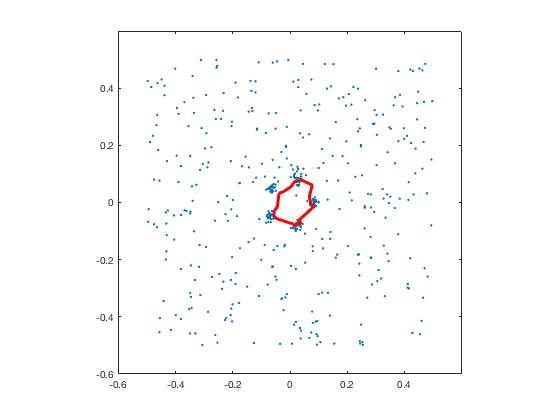}
		\caption{}
	\end{subfigure}
	\begin{subfigure}{0.3\textwidth}
		\includegraphics[width=\textwidth]{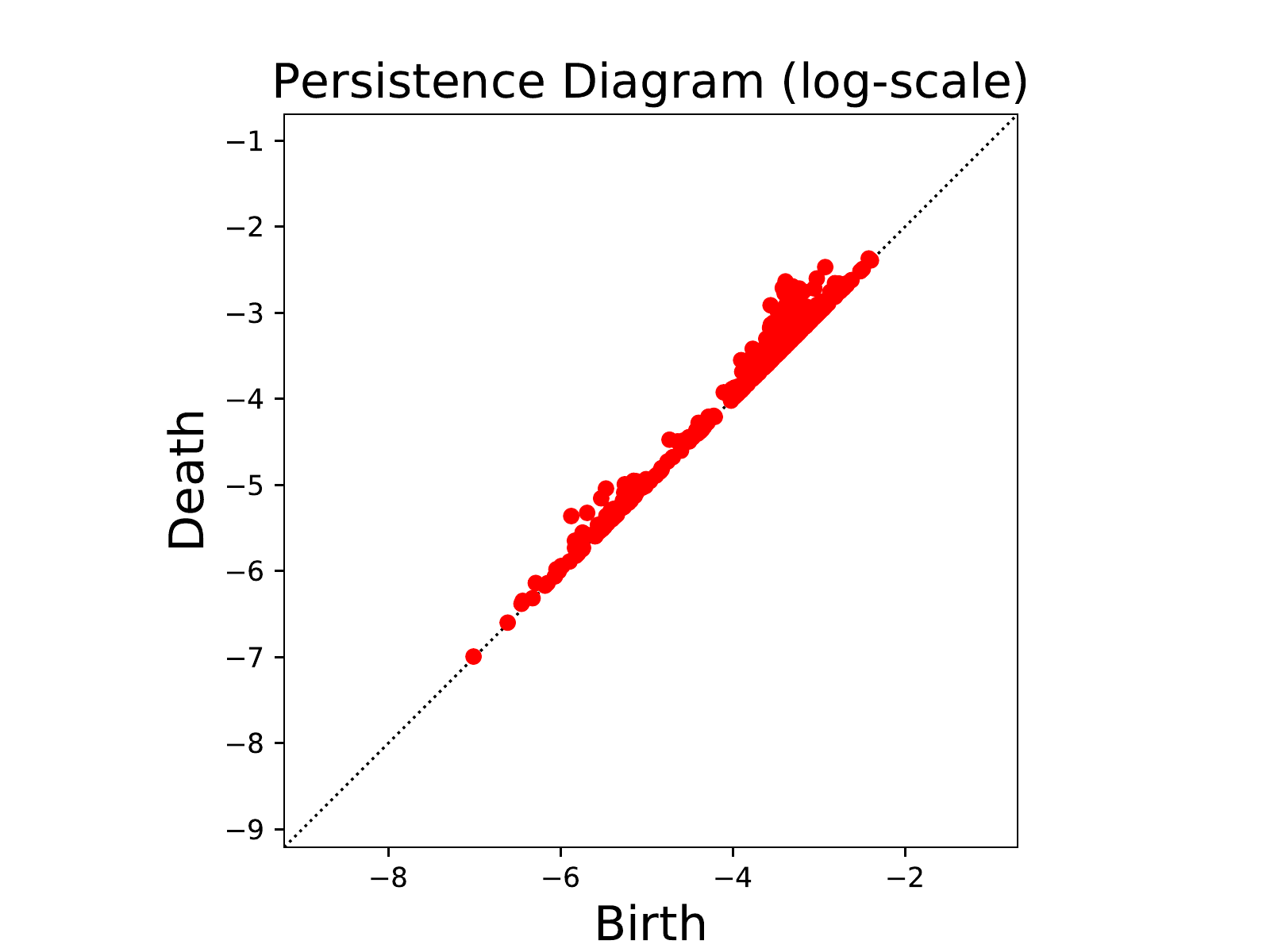}
		\caption{}
	\end{subfigure}
	\begin{subfigure}{0.3\textwidth}
		\includegraphics[width=\textwidth]{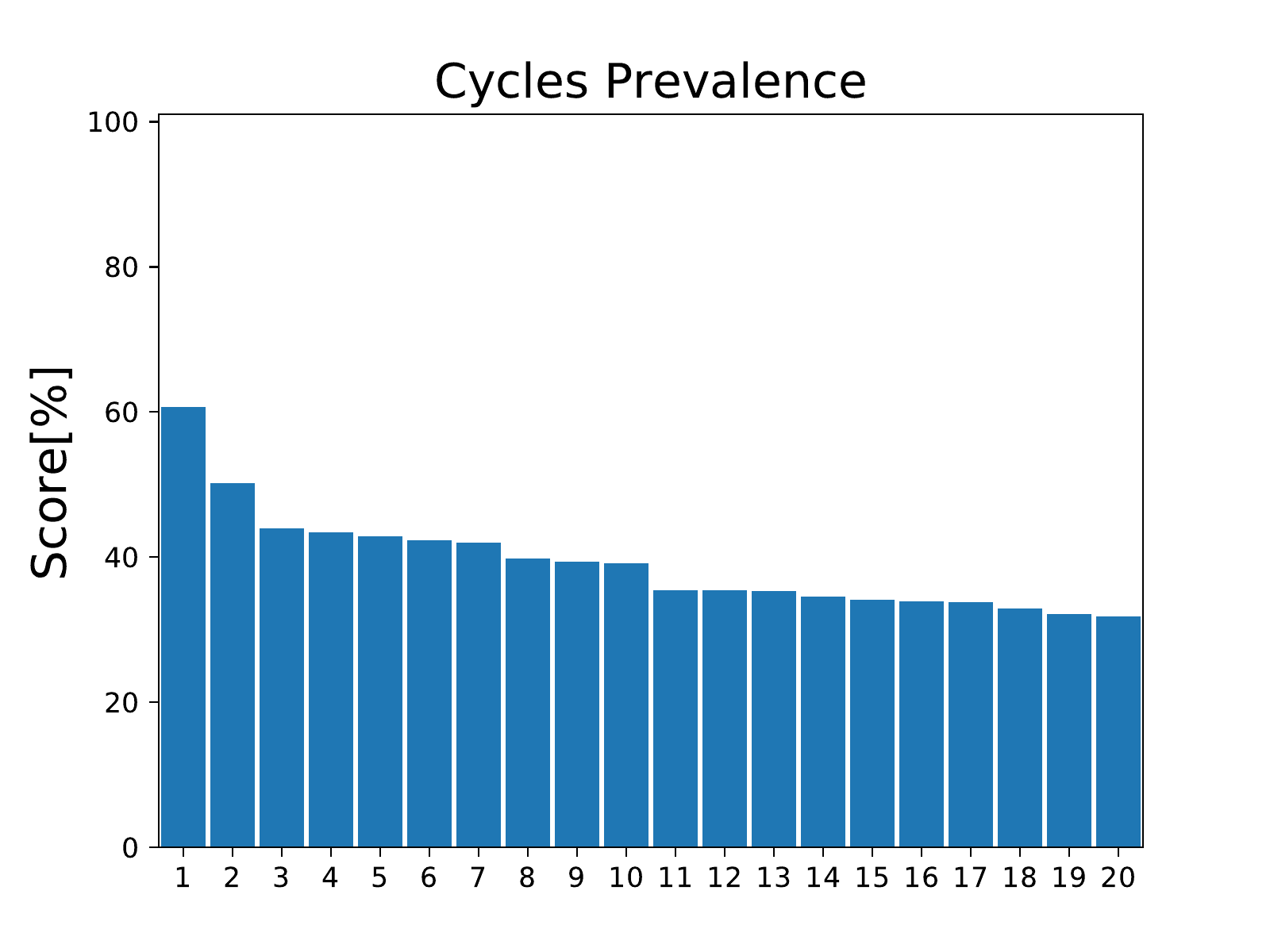}
		\caption{}
	\end{subfigure}
		\caption{Clusters on a noisy background for $p=0.3$ (first row) and $p=0.7$ (second row). (a) and (d) Original samples each with the generator (in red) with the highest cycle prevalence. (b) and (e) Persistence diagrams (log scale). In both diagrams there are no prominent cycles. (c) and (f) Cycle prevalence. As can be seen in both cases the cycle prevalence detects the cycle supported on the clusters. However, for $p=0.3$ it is much more prominent.}
	\label{fig:clusters}
\end{figure}

\paragraph{Clusters and noise.} In this example we generate five clusters that are masked by uniform noise. We do so by sampling $500$ points from the following random variable,
$$X = SW+(1-S)V,$$
where $W\sim U([0,1]^2)$, $S\sim \text{Bernoulli}(p)$  and $V$ is the following mixture of Gaussians
$$f_V(v)=\dfrac{1}{5}\sum_{i=1}^5 \dfrac{1}{\sqrt{2\pi \sigma^2}}e^{\dfrac{-\|v-\mu_i\|^2}{2\sigma^2}},$$
where $\mu_i = 0.1\big( \cos\big(\frac{2\pi}{5}(i-1) \big), \sin\big(\frac{2\pi}{5}(i-1)\big)\big)$. 
We generate additional $100$ resamples, each contains $500$ points from the kernel density estimator with $h=10^{-6}$. The idea behind this example is that when we resample the data we are likely to get several points from each cluster. Hence, the cycle induced by the clusters is likely to reappear in many resamples. However, the persistence of this cycle is of the same magnitude of the noisy cycles that appear in the data, since the distance between adjacent clusters puts a bound on its birth value, which makes it undetectable in the persistence diagram. 
Figure \ref{fig:clusters} presents the results for both $p=0.3$ and  $p=0.7$. 
As can be seen in the figures, although the cycle induced by the clusters is undetectable in both persistence diagrams, its cycle prevalence is the highest in both experiments. However, the percentage of noise has major effect on its score, for $p=0.7$ the score drops significantly compared to $p=0.3$. 

\begin{figure}[h]
\centering
	\begin{subfigure}{0.24\textwidth}
		\includegraphics[width=\textwidth]{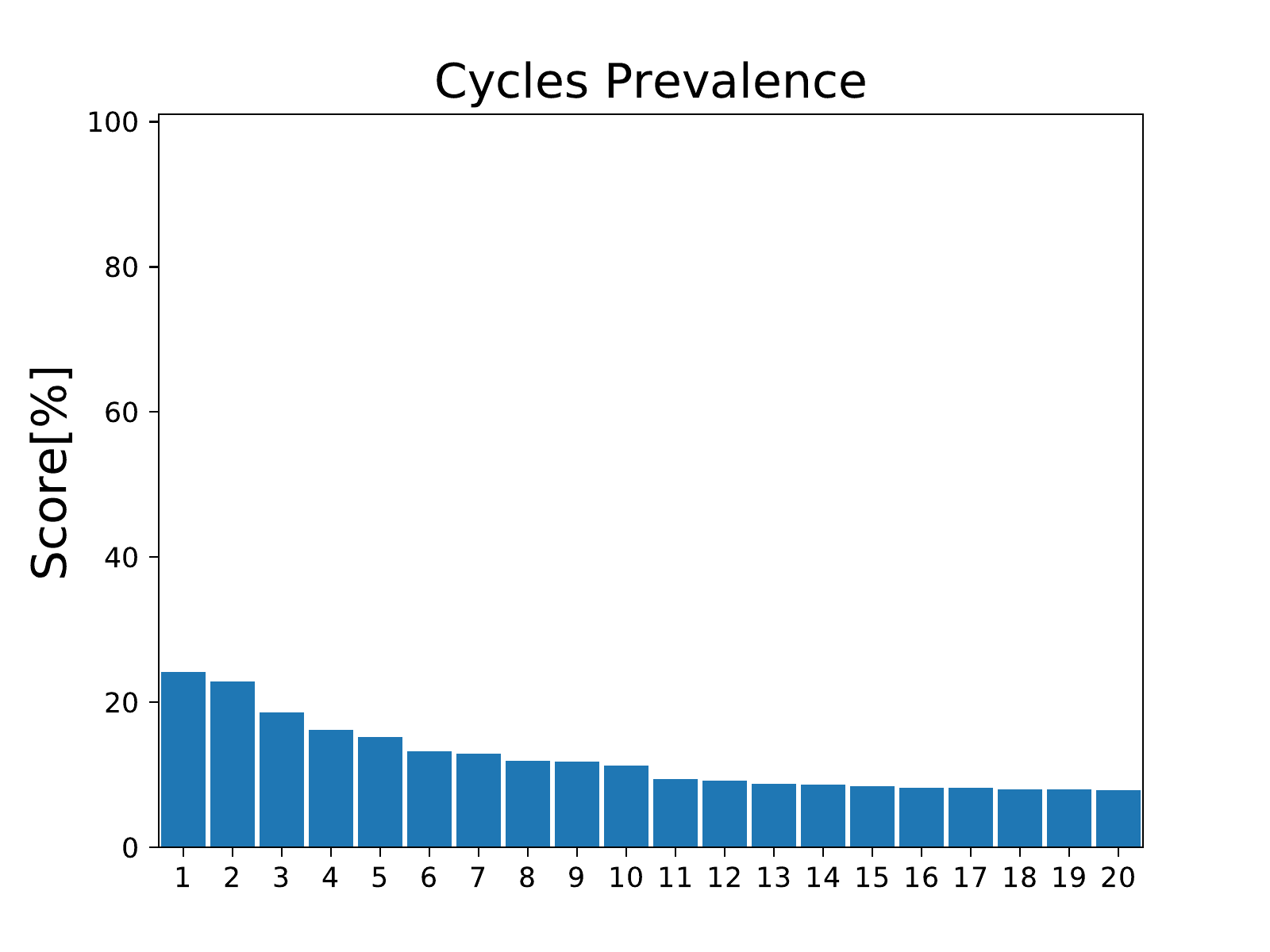}
		\caption{}
	\end{subfigure}
	\begin{subfigure}{0.24\textwidth}
		\includegraphics[width=\textwidth]{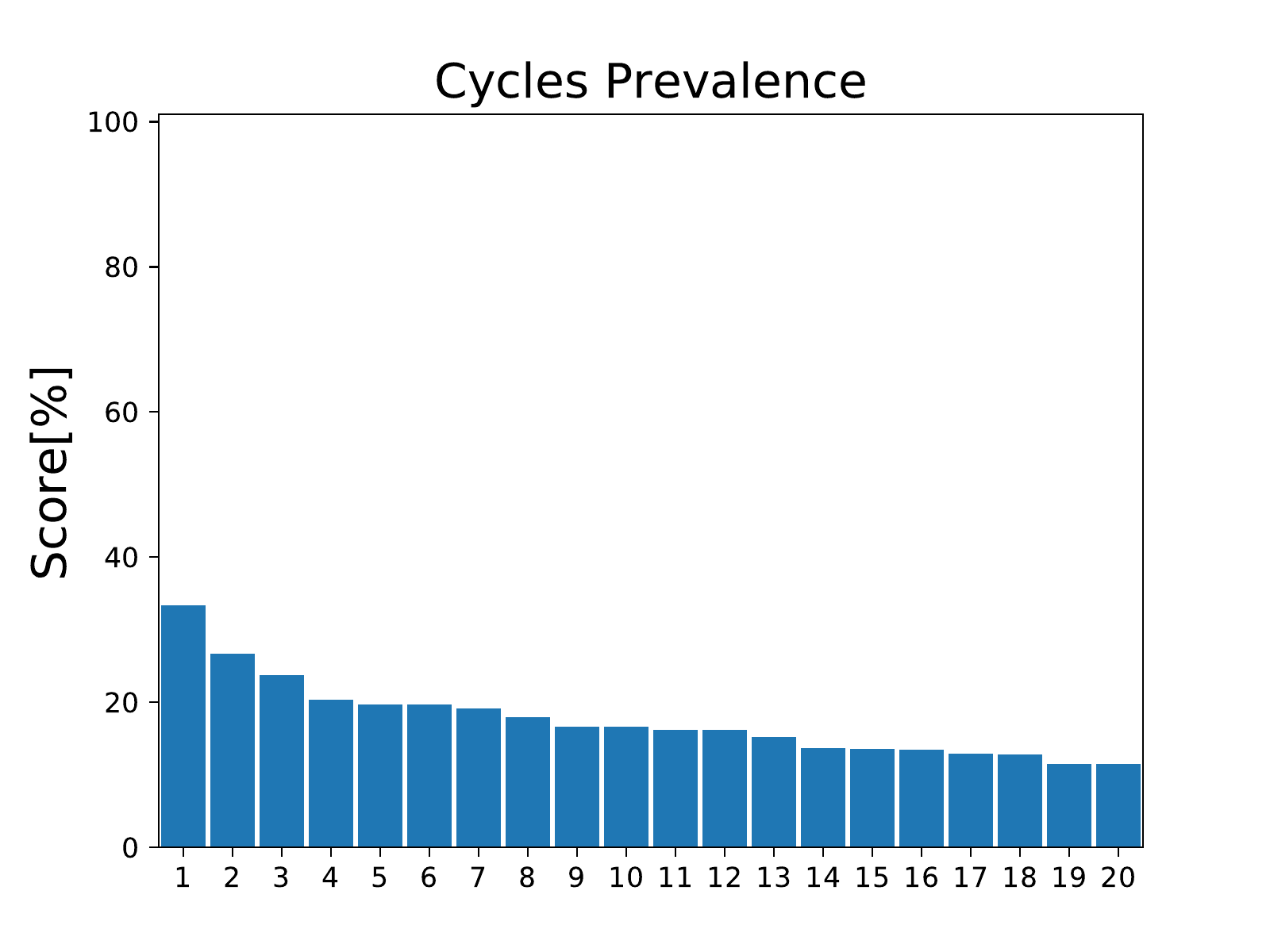}
		\caption{}
	\end{subfigure}
	\begin{subfigure}{0.24\textwidth}
		\includegraphics[width=\textwidth]{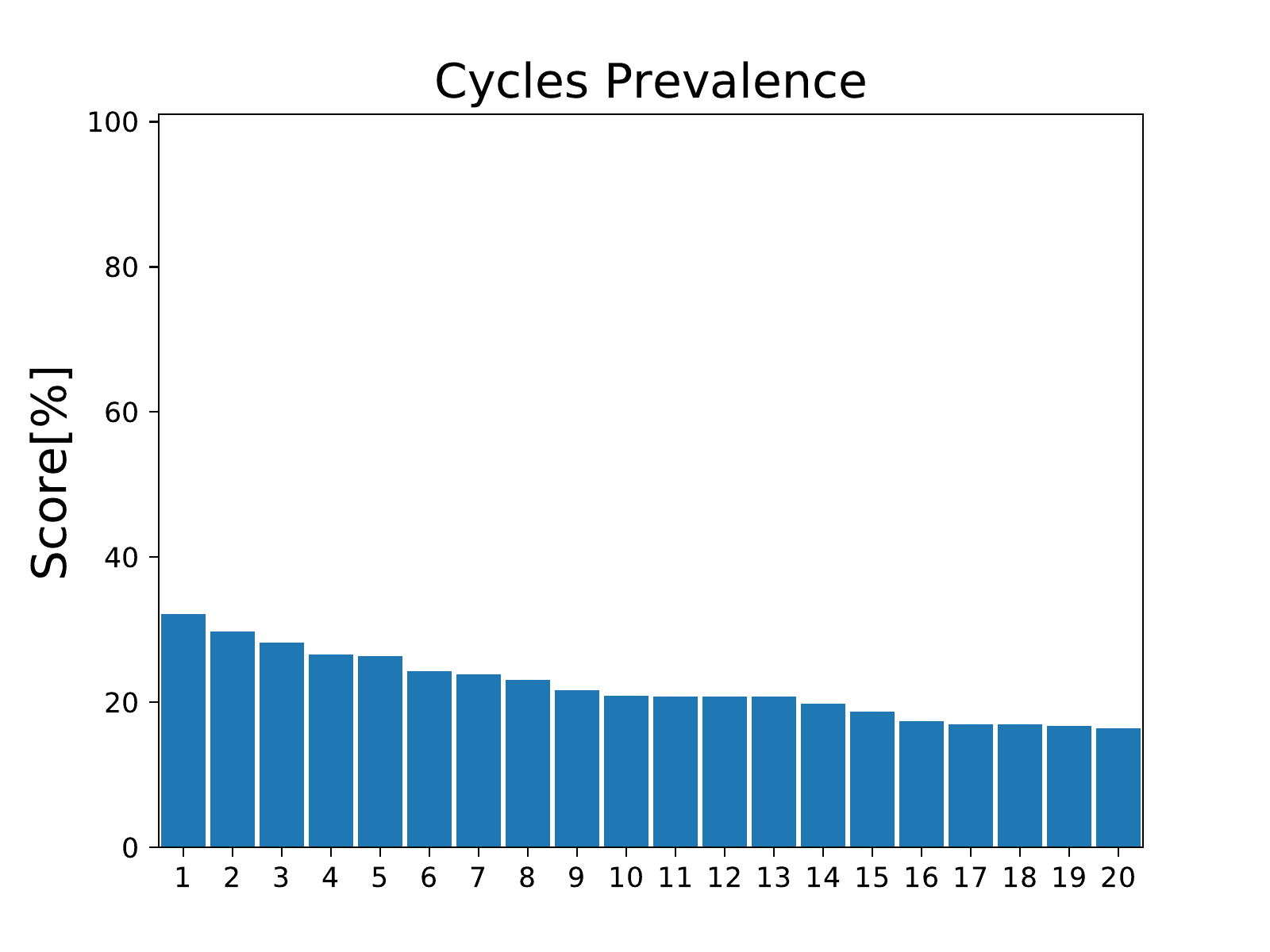}
		\caption{}
	\end{subfigure}
	\begin{subfigure}{0.24\textwidth}
		\includegraphics[width=\textwidth]{graphics/chapter3/noise/match_example_5.png}
		\caption{}
		\label{fig:large_noise}
	\end{subfigure}
	\caption{Cycle prevalence scores for points sampled uniformly from $[0,1]^2$. (a) $n=500$, (b) $n=1000$ and (c) $n=2000$. (d) Matching cycles example for $n=1000$. Red: original cycle. Blue: matching cycle.}
	\label{fig:noise}
\end{figure}

\paragraph{Noise.} As mentioned above, while experimenting with simulated data we found that noisy cycles tend to reappear in many samples. Hence, measures that put too much weight on frequency lose their effectiveness when the sample size increase (see Figure \ref{fig:noise_problem}).
The following example suggests that the prevalence  of noisy cycles is bounded regardless of sample size, a property which ensures that we are able to separate signal that stands out from noise effectively for large samples as well. We ran this test for points in $\mathbb{R}^2$, for $n=500,1000,2000$ and $B=100$. the results are shown in Figure \ref{fig:noise}. 

\subsection{A new metric for persistence diagrams}\label{sec:metric}
As we stated earlier, the literature in applied topology contains several notions of metrics to compare between persistence diagrams. One of the most commonly used is the Wasserstein distance, which is defined as follows. Suppose that $\PH_1$ and $\PH_2$ are persistence modules, and let $\PD_1,\PD_2$ be their corresponding persistence diagrams, considered as finite subsets of $\R^2$.  The $p$-th Wasserstein distance is defined as follows.
\[
d_{W_p}(\PD_1,\PD_2) := \inf_{\phi:\widehat\PD_1\to\widehat\PD_2} \left( \sum_{x\in \widehat\PD_1} \|x-\phi(x)\|^p\right)^{1/p},
\]
where $\widehat \PD_1,\widehat \PD_2$ are augmented versions of $\PD_1,\PD_2$ that include diagonal points, and $\phi$ goes over all possible bijections (see \cite{skraba2020wasserstein}), and where $\|\cdot\|$ is commonly taken to be the supremum norm.

The Wasserstein distance has been proven to be a powerful metric for persistence diagrams. However, as we mentioned in the introduction, it only considers the numerical values of the persistence diagram rather than the entire algebraic structure of  persistent homology. This can become a significant drawback when completely different spaces may produce similar persistence diagrams (see Figure \ref{fig:diff_spaces_same_hom}).

\begin{figure}[h]
	\centering
\includegraphics[width=0.8\textwidth]{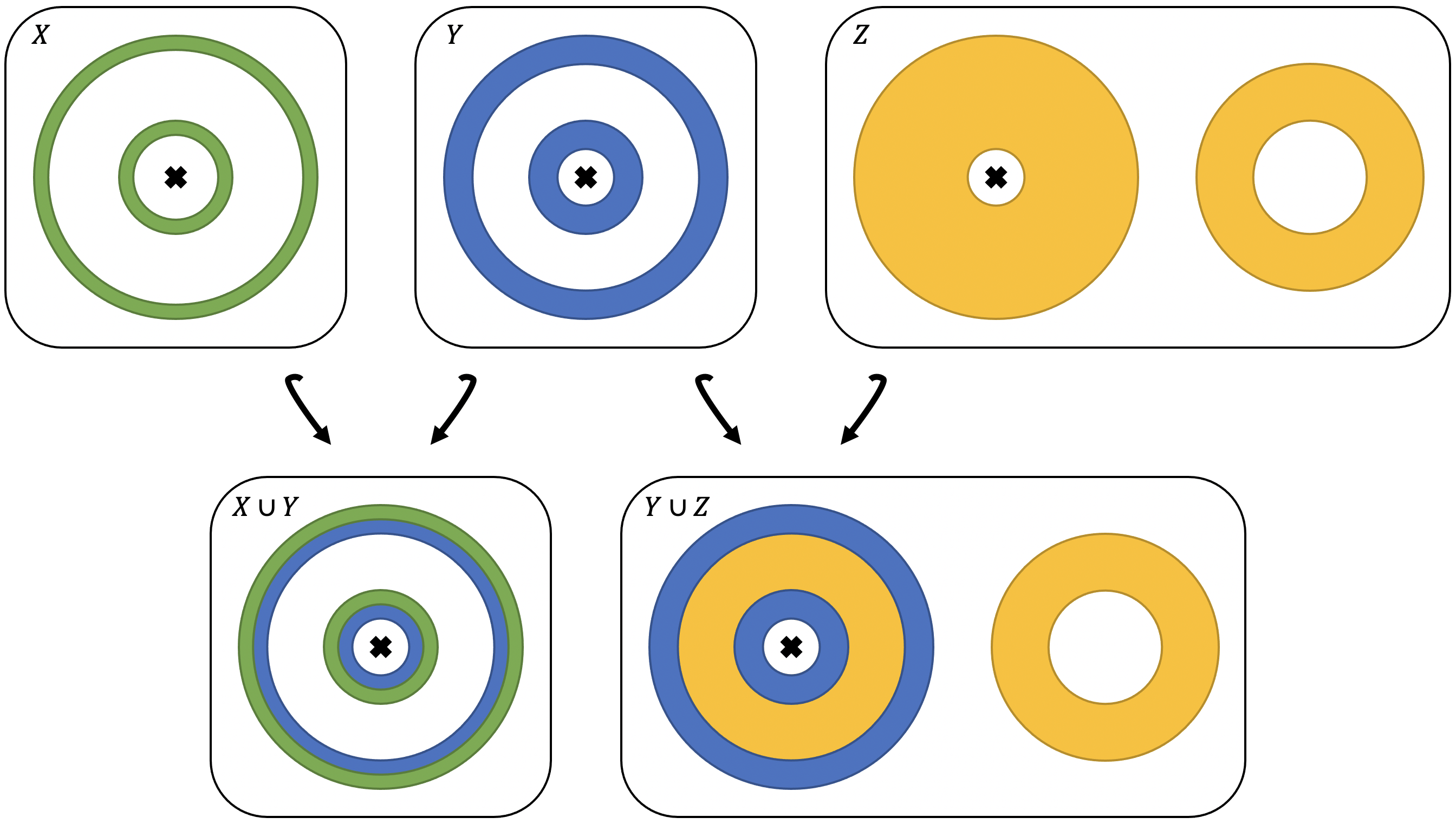}
	\caption{Wasserstein-like distance example. 
	Consider the spaces $X,Y,Z \subset \R^2$ above, where `x' marks the origin, and the largest circle is of radius $1$. We want to compute distances between these shapes based on $\PH_1$ for the filtration generated by the union of balls around each space. In this case, all cycles have birth time = $0$, and the persistence diagrams are
	$\PD_1(X) = \{(0,0.3),(0,0.5)\}$, $\PD_1(Y) = \{(0,0.2),(0,0.4)\}$, and $\PD_1(Z) = \{(0,0.2),(0,0.4)\}$. Consequently, we have $d_{W_1}(\PD_1(Y),\PD_1(X)) = 0.2$, and $d_{W_1}(\PD_1(Y),\PD_1(Z)) = 0$.
	Clearly, these distances do not correctly reflect the similarity between the spaces. On the other hand, we can compute our new distance by first matching cycles based on the inclusion maps in the figure. In this case we have
	$d_{\mathrm{IM}_1}(\PH_1(Y),\PH_1(X)) = 0.2$ and $d_{\mathrm{IM}_1}(\PH_1(Y),\PH_1(Z))=0.8$, which agrees with our intuition.}
	\label{fig:wass_like_ex}
\end{figure}

Inspired by the Wasserstein distance, we propose here a new distance measure that is based on our interval matching in Definition \ref{defn:match_pd}. Suppose that we have $X,Y,Z,f,g$ as in the setting of Section \ref{sec:ph_match}. We want to compute a distance between $\PH_k(X)$ and $\PH_k(Y)$ (rather than the persistence diagrams). For every persistence interval $\gamma$ we define $\pi(\gamma) = (\bth(\gamma),\dth(\gamma))$, i.e.~the corresponding point in the persistence diagram. In addition, for every $\gamma$ we define $\pi_0(\gamma)$
to be the nearest point to $\pi(\gamma)$ that lies on the diagonal $x=y$. In the following, $\gamma$ will denote a persistence interval in $\PH_k(X)$, and $\delta$ will denote an interval in $\PH_k(Y)$. We will use $\gamma\sim\delta$ as a shorthand notation for $\gamma\overset{\mathcal{F}_{Z,\Int}}{\sim}\delta$. We also use $\gamma\sim\emptyset$ to denote that $\gamma\in \PH_k(X)$ has no match in $\PH_k(Y)$ (and similarly for $\delta$). With this notation, we can define the following distance
\[
d_{\mathrm{IM}_p}(\PH_k(X),\PH_k(Y)) := \left(\sum_{\gamma\sim\delta} \|\pi(\gamma)-\pi(\delta)\|^p + \sum_{\gamma\sim\emptyset} \|\pi(\gamma)-\pi_0(\gamma)\|^p + \sum_{\delta\sim\emptyset} \|\pi(\delta)-\pi_0(\delta)\|^p\right)^{1/p}.
\]
In other words, similarly to the Wasserstein distance, we generate a bijection between the points in the (augmented) persistence diagrams and compute the $L^p$ distance between them. However, there are two significant differences between the two metrics. In our matching-based distance, points in the persistence diagrams are matched only if the underlying persistence interval match at the persistent homology level. This way we avoid comparing two features that have similar birth/death values but represent completely different topological phenomena. See Figure \ref{fig:wass_like_ex} for an example.
In addition, to compute the Wasserstein distance we are required to optimize the distance over all possible bijections. However, since the matching in Definition \ref{defn:match_pd} is unique, our matching-based distance does not require such an optimization procedure.

\section{Conclusion}
\label{sec:conclusion}

In this paper we introduced a new approach for cycle registration in persistent homology, motivated by the problem of differentiating between real features and noise in persistent homology. 
The main difference between our approach and previous work (e.g.~\cite{chazal_robust_2017,fasy_confidence_2014}), is that we analyze the topology of the data directly, rather than rely on low-dimensional numerical summaries. We presented the new cycle-registration framework in a generic way that  applies to various types of spaces. We focused on the case of simplicial complexes to present detailed algorithms and simulated examples. In the context of homological inference, using our framework allows us to develop bootstrap-like methods to evaluate a special kind of statistical significance measures that are able to detect the topological signal (see Section \ref{sec:hom_inf}). In the context of machine learning, our interval matching scheme can replace the bijection between persistence diagrams  in order to define Wasserstein-like metrics (see Section \ref{sec:metric}).
We consider this work as  laying the foundations for this new perspective in statistical TDA, leaving much room for future work. 

In the simulated data examples presented in Section \ref{sec:hom_inf}, the real features indeed stand out in the cycle prevalence bootstrap results, demonstrating the potential of our framework. We believe that with some further analysis, this  framework will  provide statistical tools for topological inference and other tasks.
For example, one important statistical task will be to analyze the prevalence of cycles in a suitable null distribution. Doing so will enable us to choose prevalence threshold values based on calculated p-values.
In addition, we note that the prevalence measure we defined here, is merely an example for a statistic that one can calculate from the resampled data.  
An important future work is to search for other meaningful statistics that may either detect signal better, or emphasize certain patterns in given data.

The new metric we introduced in Section \ref{sec:metric} can become a powerful and robust tool for comparing between persistence diagrams. 
As future work, we should study in detail its properties and performance, especially with respect to stability. In addition, our current definition of this metric might render it sensitive to translations and rotations,  which in some cases is undesired. A possible solution would be to include an optimization step to the distance calculation that finds the best cycle-matching among all possible rigid motions.

In addition to the statistical applications, the suggested cycle registration scheme can be used in any application where a frame of reference is available. The following are examples for such applications which we plan to pursue in future work.

Recall that in the fixed homology setting we assumed $X\xrightarrow{f} Z \xleftarrow{g} Y$. 
In that case, our cycle registration method can be viewed as computing a $1$-step zigzag persistence \cite{carlsson_zigzag_2010}. Briefly, in the simplicial setting, zigzag persistence encodes the changes to homology  over a sequence of additions and deletions of simplexes (rather than just addition as in persistent homology). For instance, the image-persistent cycles in $X$ that have a matching cycle in $Y$ correspond to cycles that persist through the single-step zigzag $X\rightarrow Z \leftarrow Y$.
This insight may lead to the development of algorithms for computing zigzag persistence based on image-persistent homology. The main challenges here are: (1) expand the above idea from a single zigzag step to a sequence, and  (2) investigate whether this approach achieves better computation times than existing algorithms \cite{milosavljevic_2011_zigzag}.

The notion of cycle registration is not limited to point cloud data and can be applied to other types of data as well.
One example could be medical imaging  (e.g.~the CT example presented in the introduction) where every two slices have a natural common frame of reference which is the 3D reconstruction of the scanned area. In that case, one can study the relations between two slices, for example the trajectories of blood vessels from one slice to another.
This requires expanding our ideas from data points into image data and functions.

%
%

\newpage

\vskip 0.2in
\bibliographystyle{plain}
\bibliography{general}

\end{document}